\documentclass[10pt,journal,compsoc]{IEEEtran}

\ifCLASSOPTIONcompsoc
  \usepackage[nocompress]{cite}
\else
  \usepackage{cite}
\fi

\ifCLASSINFOpdf
\else
\fi

\usepackage{graphicx}
\usepackage{epsfig}
\usepackage{amsmath}
\usepackage{amssymb}
\usepackage{color}
\usepackage{multirow}
\usepackage{makecell}
\usepackage{subcaption}
\usepackage{sidecap}
\usepackage[table]{xcolor}
\usepackage[group-separator={,}]{siunitx}
\usepackage{colortbl,hhline}
\usepackage{enumitem}
\usepackage{stmaryrd}
\usepackage{mathrsfs}
\usepackage{hyperref}

\definecolor{gray1}{rgb}{0.84,0.84,0.84}
\definecolor{gray2}{rgb}{1,0.89,0.75}
\definecolor{clr3}{rgb}{0.95,0.95,0.95}
\definecolor{clr4}{rgb}{0.96,0.96,0.86}

\def\method#1{{\small{\texttt{#1}}}}

\begin{document}

\title{Polarity Loss for Zero-shot Object Detection}

\author{Shafin Rahman, 
        Salman Khan 
        and Nick Barnes
\thanks{S. Rahman is with  the  Research  School  of  Engineering,  The  Australian National  University,  Canberra,  ACT 2601,  Australia,  and  also  with  Data61, Commonwealth Scientific and Industrial Research Organization,  Canberra, ACT 2601, Australia (e-mail: shafin.rahman@anu.edu.au).}
\thanks{S.  Khan  is  with  the  Research   School  of  Engineering,   The  Australian National University, Canberra, ACT 2601, Australia, and also with the Inception Institute of Artificial Intelligence, Abu Dhabi 0000, UAE.}
\thanks{N. Barnes is with  the  Research  School  of  Engineering,  The  Australian National  University,  Canberra,  ACT 2601,  Australia}
}


\markboth{Journal of \LaTeX\ Class Files,~Vol.~14, No.~8, August~2015}%
{Shell \MakeLowercase{\textit{et al.}}: Bare Demo of IEEEtran.cls for Computer Society Journals}
\IEEEtitleabstractindextext{%
\begin{abstract}

Conventional object detection models require large amounts of training data. In comparison, humans can recognize previously unseen objects by merely knowing their semantic description. To mimic similar behaviour, zero-shot object detection aims to recognize and localize `\emph{unseen}’ object instances by using only their semantic information. The model is first trained to learn the relationships between visual and semantic domains for \emph{seen} objects, later transferring the acquired knowledge to totally \emph{unseen} objects. This setting gives rise to the need for correct alignment between visual and semantic concepts, so that the \emph{unseen} objects can be identified using only their semantic attributes. In this paper, we propose a novel loss function called  \emph{`Polarity loss'}, that promotes correct visual-semantic alignment for an improved zero-shot object detection.  On one hand, it refines the noisy semantic embeddings via metric learning on a \emph{`Semantic vocabulary'} of related concepts to establish a better synergy between visual and semantic domains. On the other hand, it explicitly maximizes the gap between positive and negative predictions to achieve better discrimination between seen, unseen and background objects.  Our approach is inspired by 
embodiment theories in cognitive science, that claim human semantic understanding to be grounded in past experiences (seen objects), related linguistic concepts (word vocabulary) and 
visual perception (seen/unseen object images). We conduct extensive evaluations on MS-COCO and Pascal VOC datasets, showing  significant improvements over state of the art. Our code and evaluation protocols available at: \url{https://github.com/salman-h-khan/PL-ZSD_Release}
\end{abstract}

}

\maketitle
\IEEEdisplaynontitleabstractindextext
\IEEEpeerreviewmaketitle

\IEEEraisesectionheading{\section{Introduction}\label{sec:introduction}}

\IEEEPARstart{E}{}xisting deep learning models generally perform well in a fully supervised setting with large amounts of annotated data available for training. Since large-scale supervision is expensive and cumbersome to obtain in many real-world scenarios, the investigation of learning under reduced supervision is an important research problem. In this pursuit, human learning provides a remarkable motivation since humans can learn with very limited supervision. Inspired by human learning, zero-shot learning (ZSL) aims to reason about objects that have never been seen before using only their semantics. A successful ZSL system can help pave the way for life-long learning machines that intelligently discover new objects and incrementally enhance their knowledge. 

Traditional ZSL literature only focuses on `\emph{recognizing}' unseen objects. Since real-world objects only appear as a part of a complete scene, the newly introduced zero-shot object detection (ZSD) task \cite{rahman2018ZSD,Bansal_2018_ECCV} considers a more practical setting where the goal is to simultaneously `\emph{locate and recognize}' unseen objects. This task offers new challenges for the existing object detection frameworks, the most important being the accurate alignment between visual and semantic concepts. If a sound alignment between the two heterogeneous domains is achieved, the unseen objects can be detected at inference using only their semantic representations. Further, a correct alignment can help the detector in differentiating between the background and the previously unseen objects in order to successfully detect the unseen objects during inference.

In this work, we propose a single-stage object detection pipeline underpinned by a novel objective function named Polarity loss. Our approach focuses on learning the complex interplay between visual and semantic domains such that the unseen objects (alongside the seen ones) can be accurately detected and localized. To this end, the proposed loss simultaneously achieves two key objectives during end-to-end training process. First, it learns a new vocabulary metric that improves the semantic embeddings to better align them with the visual concepts. 
Such a dynamic adaptation of semantic representations is important since the original unsupervised semantic embeddings  (e.g., word2vec) are noisy and thereby complicate the visual-semantic alignment. Secondly, it directly maximizes the margin between positive and negative detections to encourage the model to make confident predictions. This constraint not only improves visual-semantic alignment (through maximally separating class predictions) but also helps distinguish between background and unseen classes.

Our ZSD approach is distinct from existing state-of-the-art \cite{Bansal_2018_ECCV,Demirel_BMVC_2018,rahman2018ZSD,zhu2018zero,Li_AAAI_2019} due to its direct emphasis on achieving better visual-semantic alignment. For example, \cite{rahman2018ZSD,Bansal_2018_ECCV,Demirel_BMVC_2018,zhu2018zero} use fixed semantics derived from unsupervised learning methods that can induce noise in the embedding space. In order to overcome noise in the semantic space, \cite{Li_AAAI_2019} resorts to using detailed semantic descriptions of object classes instead of single vectors corresponding to class names. Further, the lack of explicit margins between positive and negative predictions leads to confusion between background and unseen classes \cite{Bansal_2018_ECCV,zhu2018zero,Demirel_BMVC_2018}. In summary, our main contributions are:
\begin{itemize}[noitemsep,topsep=2pt]
\item An end-to-end single-shot ZSD framework based on a novel loss function called `\emph{Polarity loss}' to address object-background imbalance and achieve maximal separation between positive and negative predictions.
\item {Using an external vocabulary of words, our approach learns to associate semantic concepts with both seen and unseen objects. This helps to resolve confusion between unseen classes and background, and to appropriately reshape the noisy word embeddings.
}
\item A new seen-unseen split on the MS-COCO dataset that respects practical considerations such as diversity and rarity among unseen classes. 
\item Extensive experiments on the old and new splits for MS-COCO and Pascal VOC which give absolute gains of $9.3$ and $7.6$ in mAP over \cite{Bansal_2018_ECCV} and  \cite{Demirel_BMVC_2018}, respectively. 
\end{itemize}

A preliminary version of this work appeared in~\cite{Rahman_AAAI_2020}, where the contribution is limited to a single formulation of the polarity loss. Moreover, the experiments only consider word2vec as word embedding. 
In the current paper, we extended our work as follows: (1) An alternate formulation of Polarity loss (Sec.~\ref{sec:ourloss}). (2) Motivation of the proposed new split based on MSCOCO dataset for zero-shot detection (Sec.~\ref{sec:exp}). (3) A detailed discussion on related works in the literature (Sec.~\ref{sec:related_work}). (4) More experiments and ablation studies e.g., with multiple semantic embeddings including Word2vec, GloVe and FastText, a validation study on hyper-parameters and studying the impact of changing overlap threshold (Sec.~\ref{sec:exp}).

\section{Related work} \label{sec:related_work}

 \textbf{Zero-shot learning (ZSL):} The earliest efforts were based on manually annotated attributes as a mid-level semantic representation \cite{Lampert_PAMI_2014}. This approach resulted in a decent performance on fine-grained recognition tasks but to eliminate strong attribute supervision; researchers start exploring unsupervised word-vector based techniques \cite{Mikolov_NIPS_2013,Jeffrey_Glove_2014}. {Independent of the source of semantic information, a typical ZSL method needs to map both visual and semantic features to a common space to properly align information available in both domains. This can be achieved in three ways: (a) transform the image feature to semantic feature space \cite{Rahman_TIP_2018}, (b) map the semantic feature to image feature space \cite{Zhang_2017_CVPR,Kodirov_2017_CVPR} or, (c) map both image or semantic features to a common intermediate latent space \cite{Xian_2016_CVPR,Yu_Latent_2018}.} 
 To apply ZSL in practice, a few notable problem settings are: \textbf{(1)} transductive ZSL \cite{Song_2018_CVPR,Yu_Transductive_2018}: making use of unlabeled unseen data during training, \textbf{(2)} generalized ZSL \cite{Felix_2018_ECCV,Rahman_TIP_2018}: classifying seen and unseen classes together, \textbf{(3)} domain adaptation \cite{Kodirov_2015_ICCV,Zhao_NIPS_2018}: learning a projection function to adapt unseen target to seen source domain, \textbf{(4)} class-attribute association \cite{Al-Halah_2016_CVPR,al2017automatic}: relating unsupervised semantics to human recognizable attributes. In this paper, our focus is not only recognition but also simultaneous localization of unseen objects which is a significantly complex and challenging problem.
 
  \begin{figure}[t]
  \centering
   \includegraphics[width=1\columnwidth,trim={0cm 0.7cm 0cm 0cm},clip]{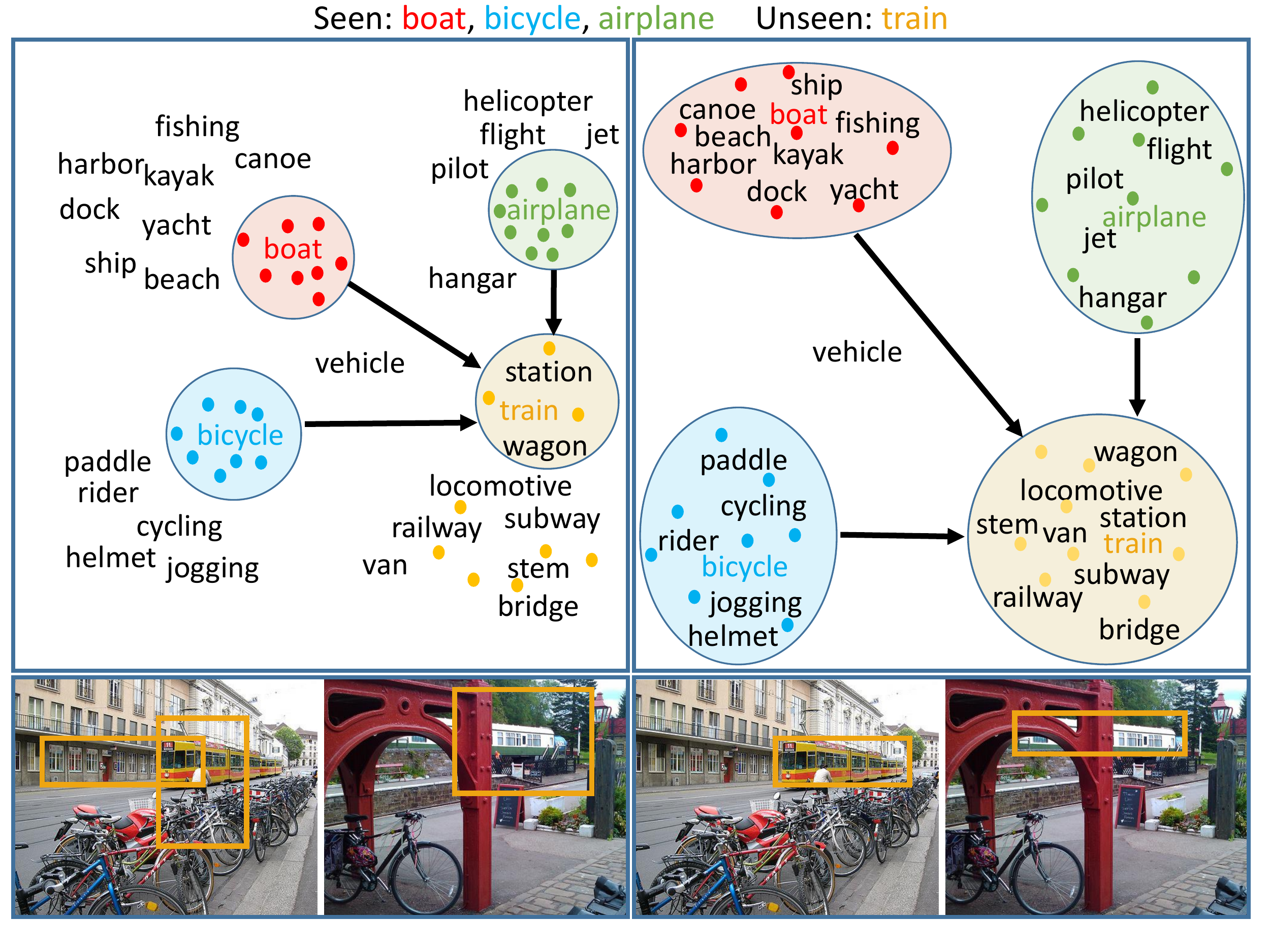}
   \caption{\small (\emph{Top Left}) Traditional ZSD approaches align visual features (solid dots) to their corresponding semantics (e.g., boat, airplane) without considering the related semantic concepts (black text). It results in a fragile description of an unseen class (train) and causes confusion with background and seen classes ({bottom left}). (\emph{Top Right}) Our approach automatically attends to related semantics from an external vocabulary and reshapes the semantic embedding so that visual features are well-aligned with 
   related semantics. Moreover, it maximizes the inter-class separation that avoids confusion between unseen and background (bottom right).}
\label{fig:overview_polar}
\end{figure}
 
\textbf{Object detection:} End-to-end trainable deep learning models have set new performance records on object detection benchmarks.  The most popular deep architectures can be categorized into double stage networks (e.g., FasterRCNN~\cite{Faster_RCNN_2017}, RFCN~\cite{Dai_RFCN_2016}) and single stage networks (e.g., SSD\cite{Liu_SSD_2016}, YOLO\cite{Redmon_yolo9000_2016}). Generally, double-stage detectors achieve high accuracy, while single-stage detectors work faster. To further improve performance, recent object detectors introduce novel concepts such as feature pyramid network (FPN)~\cite{FPN_2017_CVPR,Kong_2018_ECCV} instead of region proposal network (RPN), focal loss~\cite{RetinaNet_2017_ICCV} instead of traditional cross-entropy loss, non-rectangular region selection~\cite{Xu_2018_ECCV} instead of rectangular bounding boxes, designing backbone architectures~\cite{Li_2018_ECCV} instead of ImageNet pre-trained networks (e.g., VGG\cite{Vgg_arXiv_2014}/ResNet\cite{ResNet_CVPR_2016}). In this paper, we attempt to extend object detection to the next level: detecting unseen objects which are not observed during training.

  \textbf{Zero-shot object detection (ZSD):} 
The ZSL literature is predominated by classification approaches that focus on single \cite{Fu_survey_2018,Xian_PAMI_2018,Lampert_PAMI_2014,Rahman_TIP_2018,Zhang_2017_CVPR,Kodirov_2017_CVPR} or multi-label~\cite{Lee_2018_CVPR,rahman2018deep} recognition. The extension of conventional ZSL approaches to zero-shot object localization/detection is relatively less investigated. Among previous attempts, Li \emph{et al.} \cite{li2014attributes} learned to segment attribute locations which can locate unseen classes. \cite{jetley2016straight} used a shared shape space to segment novel objects that look like seen objects. These approaches are useful for classification but not extendable to ZSD. \cite{hu2016natural,li2017tracking} proposed novel object localization based on natural language description. Few other methods located unseen objects with weak image-level labels \cite{rahman2018deep,Ren_BMVC_17}. However, none of them perform ZSD. Very recently, \cite{rahman2018ZSD,zhu2018zero,Bansal_2018_ECCV,Demirel_BMVC_2018,Li_AAAI_2019} investigated the ZSD problem. These methods can detect unseen objects using box annotations of seen objects. Among them, \cite{Bansal_2018_ECCV} proposed a feature based approach where object proposals are generated by edge-box \cite{Edge_Boxes_2014}, and \cite{rahman2018ZSD,zhu2018zero,Demirel_BMVC_2018} modified the object detection frameworks~\cite{Faster_RCNN_2017,Redmon_yolo9000_2016} to adapt ZSD settings. In another work, \cite{Li_AAAI_2019} use textual description of seen/unseen classes to obtain semantic representations that are used for zero-shot object detection.
Here, we propose a new loss formulation that can greatly benefit single stage zero-shot object detectors.

\textbf{Generalized Zero-shot object detection (GZSD):} This setting aims to detect both seen and unseen objects during inference. The key difference from generalized zero-shot recognition is that  multiple objects can co-exist in a single image, thereby posing a challenge for the detector. In contrast, in the recognition setting either a seen or an unseen label is assigned since a single category is assumed per image.  Only a handful of previous works in the literature reported GZSD results~\cite{Bansal_2018_ECCV,Demirel_BMVC_2019,Rahman_2019_ICCV}. \cite{Rahman_2019_ICCV} and \cite{Demirel_BMVC_2019} reported GZSD results while addressing transductive zero-shot detection and unseen object captioning problems, respectively. In the current work, we compare GZSD results with \cite{Bansal_2018_ECCV} who evaluate on GZSD problem.

\section{Polarity Loss}
We first introduce the proposed Polarity Loss that builds upon Focal
loss \cite{RetinaNet_2017_ICCV} for generic object detection. Focal loss only promotes correct prediction, whereas a sound ZSL system should also learn to minimize projections on representative vectors for negative classes. Our proposed loss jointly maximizes projection on correct classes and minimizes the alignment with incorrect ones (Sec.~\ref{sec:ourloss}). Furthermore, it allows reshaping the noisy class semantics using a metric learning approach (Sec.~\ref{sec:vocab_met}).  This approach effectively allows distinction between background \emph{vs.} unseen classes and promotes better overall alignment between visual and semantic concepts. Below in Sec.~\ref{sec:background}, we provide a brief background followed by a description of proposed loss. 

\subsection{Balanced Cross-Entroy vs. Focal loss}\label{sec:background}
Consider a binary classification task where $y \in \{0, 1\}$ denotes the ground-truth class and $p \in [0,1]$  is the prediction probability for the positive class (i.e., $y=1$). The standard binary cross-entropy (CE) formulation gives:
\begin{align}\label{eq:CE}
  \text{CE}(p,y) = -\alpha_t \log p_t, \quad p_t =\begin{cases}
    								p, & \text{if $y=1$,}\\
    								1-p, & \text{otherwise}.
 \end{cases}
\end{align}
where, $\alpha$ is a loss hyper-parameter representing inverse class frequency and the definition of $\alpha_t$ is analogous to $p_t$. In the object detection case, the object vs. background ratio is significantly high (e.g., $10^{-3}$). Using a weight factor $\alpha$ is a traditional way to address this strong imbalance. However, being independent of the model's prediction, this approach treats both well-classified (easy) and poorly-classified (hard) cases equally. It favors easily classified examples to dominate the gradient and fails to differentiate between easy and hard examples. To address this problem, Lin \emph{et al.} \cite{RetinaNet_2017_ICCV} proposed `Focal loss' (FL):
\begin{align}\label{eq:FL}
  \text{FL}(p,y) = -\alpha_t (1-p_t)^\gamma\log p_t,
\end{align}
where, $\gamma \in [0,5]$ is a loss hyper-parameter that dictates the slope of cross entropy loss (a large value denotes higher slope). The term $(1-p_t)^\gamma$ enforces a high and low penalty for hard and easy examples respectively. In this way, FL simultaneously addresses object vs. background imbalance and easy vs. hard examples difference during training. 
    
  \textbf{Shortcomings:} In zero-shot learning, it is highly important to align visual features with semantic word vectors. This alignment requires the training procedure to (1) push visual features close to their ground-truth embedding vector and (2) push them away from all negative class vectors. FL can only perform (1) but cannot enforce (2) during the training of ZSD. Therefore, although FL is well-suited for traditional seen object detection, but not for the ZSD scenario.

\subsection{Max-margin Formulation}
\label{sec:ourloss}
To address the above-mentioned shortcomings, we propose a margin maximizing loss formulation that is particularly suitable for ZSD. This formulation is generalizable and can work with loss functions other than  Eqs.~\ref{eq:CE} and \ref{eq:FL}. However, for the sake of comparison with the best model, we base our analysis on the state of the art FL. 

\subsubsection{Objective Function}
 \textbf{Multi-class Loss:} Consider that a given training set $\{\mathbf{x}, \mathbf{y}\}_i$ contains $N$ examples belonging to $C$ object classes plus an additional background class.  For the multi-label prediction case, the problem is treated as a sum of individual binary cross-entropy losses where each output neuron decides whether a sample belongs to a particular object class or not. Assume,  $\mathbf{y} = \{y^i \in \{0, 1\}\} \in \mathbb{R}^{C}$ and $\mathbf{p} = \{p^i \in [0,1]\} \in \mathbb{R}^{C}$ denotes the ground-truth label and prediction vectors respectively, and the background class is denoted by $\mathbf{y} = \mathbf{0} \in \mathbb{R}^{C}$. Then, the FL for a single box proposal is:
\begin{align}
\mathscr{L} = \sum_{i} -\alpha^i_t (1-p_t^i)^{\gamma}\log p^i_t.
\end{align}

 \textbf{Polarity Loss:} Suppose, for a given bounding box feature containing an $\ell^{th}$ object class, $p^{\ell}$ represents the prediction value for the ground-truth object class, \emph{i.e.}, $y^{\ell}=1$, see Table \ref{tab:toy}. Note that $p^{\ell}=0$ for the background class (where $y^i=0 ; \forall i$). 
Ideally, we would like to maximize the predictions for ground-truth classes and simultaneously minimize prediction scores for all other classes. We propose to explicitly maximize the margin between predictions for positive and negative classes to improve the visual-semantic alignment for ZSD (see Fig.~\ref{fig:alignment_vis}). This leads to a new loss function that we term as `Polarity Loss' (PL):
\begin{align}
  \mathscr{L}_{PL} = \sum_i f_p(p^i-p^{\ell}) \text{FL}(p^i,y^i), 
\end{align}
where, $f_p$ is a monotonic penalty function. For any prediction, $p^i$ where ${\ell} {\neq} i$, the difference $p^i{-}p^{\ell}$ represents the disparity between the true class prediction and the prediction for the negative class. The loss function enforces a large negative margin to push predictions $p^i$ and $p^{\ell}$ further apart. Thus, for an object anchor case, the above objective enforces $p^{\ell} {>} p^i$, while for background case $0 {>} p^i$ \emph{i.e.}, all $p^i$'s are pushed towards zero (since $p^{\ell}{=}0$).

 \textbf{Our Penalty Function:} $f_p$ should necessarily be a `monotonically increasing' function. It offers a small penalty if the gap $p^i{-}p^{\ell}$ is low and a large penalty if the gap is high. This constraint enforces that $p^{i} < p^{\ell}$. In this paper, we implement $f_p$ with a $\beta$ parameterized sigmoid function:
\begin{align}
f_p(p^i-p^{\ell}) = \frac{1} {1+\exp(-\beta(p^i-p^{\ell}))}
\label{eq:penalty}
\end{align}
For the case when $p^i{=}p^{\ell}$, the FL part guides the loss because $f_p$ becomes a constant. We choose a sigmoid form for $f_p$ because the difference $(p^i{-}p^{\ell}) \in [-1,1]$ and $f_p$ can be bounded by $[0,1]$, similar to $\alpha_t$ or the $(1{-}p_t)$ factor of FL. Note that, it is not compulsory to stick with this particular choice of $f_p$. We also test a softplus-based function for $f_p$ in the next subsection.


  \textbf{Final Objective:} The final form of the loss is:
\begin{align}\label{eq:final_obj}
&\mathscr{L}_{PL}(\mathbf{p},\mathbf{y}) = \sum_i \frac{-\alpha^i_t (1-p^i_t)^\gamma\log p^i_t} {1+\exp(-\beta(p^i-p^{\ell}))}, \text{  where,} \notag\\
& p^i_t =\begin{cases}
    p^i, & \text{if $y^i=1$}\\
    1-p^i, & \text{otherwise}
  \end{cases}\qquad
    p^{\ell} = p^{i} \llbracket y^{i}=1 \rrbracket,
\end{align}
where, $\llbracket \cdot \rrbracket$ denotes the Iverson bracket. Later in Sec.~\ref{sec:vocab_met}, we describe our vocabulary-based metric learning approach to obtain $\mathbf{p}$ in the above Eq.~\ref{eq:final_obj}.



\begin{table}[!tp]
    \begin{subtable}{.5\columnwidth}
      \centering
        \caption{Object case: $p^{\ell} = .8$}
        \scalebox{1}{
        \begin{tabular}{|c|c|c|c|}
        	\hline
 			$p^i$ & \cellcolor{green!10} .1 & \cellcolor{red!10} .8 & \cellcolor{red!10} .9\\ \hline
            $y^i$ & \cellcolor{green!10} 0 & \cellcolor{red!10} 1 &  \cellcolor{green!10} 0 \\ \hline
            $p_t^i$ & \cellcolor{red!10} .9 & \cellcolor{red!10} .8 & \cellcolor{green!10} .1\\ \hline
            $p^i-p^{\ell}$ & \cellcolor{green!10} -.7 & \cellcolor{green!10} 0 & \cellcolor{red!10} .1\\ \hline
            loss & \cellcolor{green!10} L & \cellcolor{green!10} L & \cellcolor{red!10} H \\ \hline
        \end{tabular}}
    \end{subtable}%
    \begin{subtable}{.5\columnwidth}
      \centering
        \caption{Background case: $p^{\ell} = 0$}
        \scalebox{1}{
        \begin{tabular}{|c|c|c|c|}
        	\hline
 			$p^i$ & \cellcolor{green!10} .1 & \cellcolor{red!10} .8 & \cellcolor{red!10} .9\\ \hline
            $y^i$ & \cellcolor{green!10} 0 & \cellcolor{green!10} 0 & \cellcolor{green!10} 0 \\ \hline
            $p_t^i$ & \cellcolor{red!10} .9 & \cellcolor{green!10} .2 & \cellcolor{green!10} .1 \\ \hline
            $p^i-p^{\ell}$ & \cellcolor{green!10} .1 & \cellcolor{red!10} .8 & \cellcolor{red!10} .9\\ \hline
            loss & \cellcolor{green!10} L & \cellcolor{red!10} H & \cellcolor{red!10} H \\ \hline
        \end{tabular}}   
    \end{subtable} 
    \caption{\small A toy example. Intermediate computations for Polarity Loss are shown. Low (L) values are shown in green while High (H) values are shown in red. A mismatch between ($p^i$ and $y^i$) + a close match between ($y^{i}$ and $y^{\ell}$)  results in a high loss.}
    \label{tab:toy}
\end{table}

\begin{figure}[!tp]
  \centering
  \includegraphics[width=1\columnwidth,trim={1cm 0cm 1cm .3cm},clip]{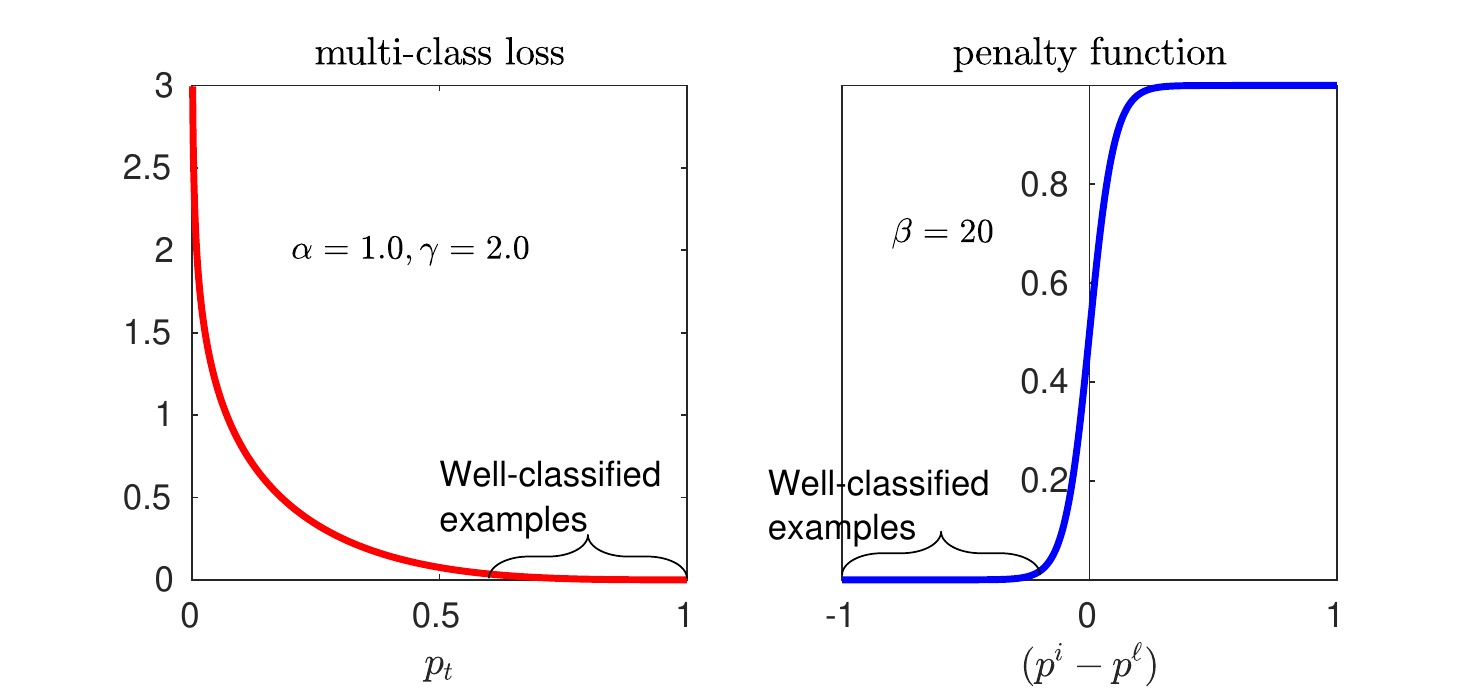}
  \caption{\small Plots of  multi-class loss (\emph{left}) and  penalty function (\emph{right}).}
\label{fig:losses}
\end{figure}

\subsubsection{Analysis and Insights}
    
 \textbf{A Toy Example:} We explain the proposed loss with a toy example in Table \ref{tab:toy} and Fig.~\ref{fig:losses}. When an anchor box belongs to an `\emph{object}' and $p_t^i \geq .5$ (high) then $p^i{-}p^{\ell} \leq 0$ (low). From Fig.~\ref{fig:losses}, both a multi-class loss and the penalty function find low loss which eventually calculates a low loss. Similarly, when $p_t^i < .5$ (low), $p^i{-}p^{\ell} > 0 $ (high), which evaluates to a high loss. When an anchor belongs to `\emph{background}', $p^i{-}p^{\ell} \geq 0$ and a high $p^i$ results in a high value for both multi-class loss and the penalty function and vice versa. In this way, the penalty function always supports multi-class loss based on the disparity between the current prediction and ground-truth class's prediction.

\begin{figure}[!tp]
  \centering
\includegraphics[width=1\columnwidth, height=3.5cm,trim={1.8cm 1.4cm 2.2cm .3cm},clip]{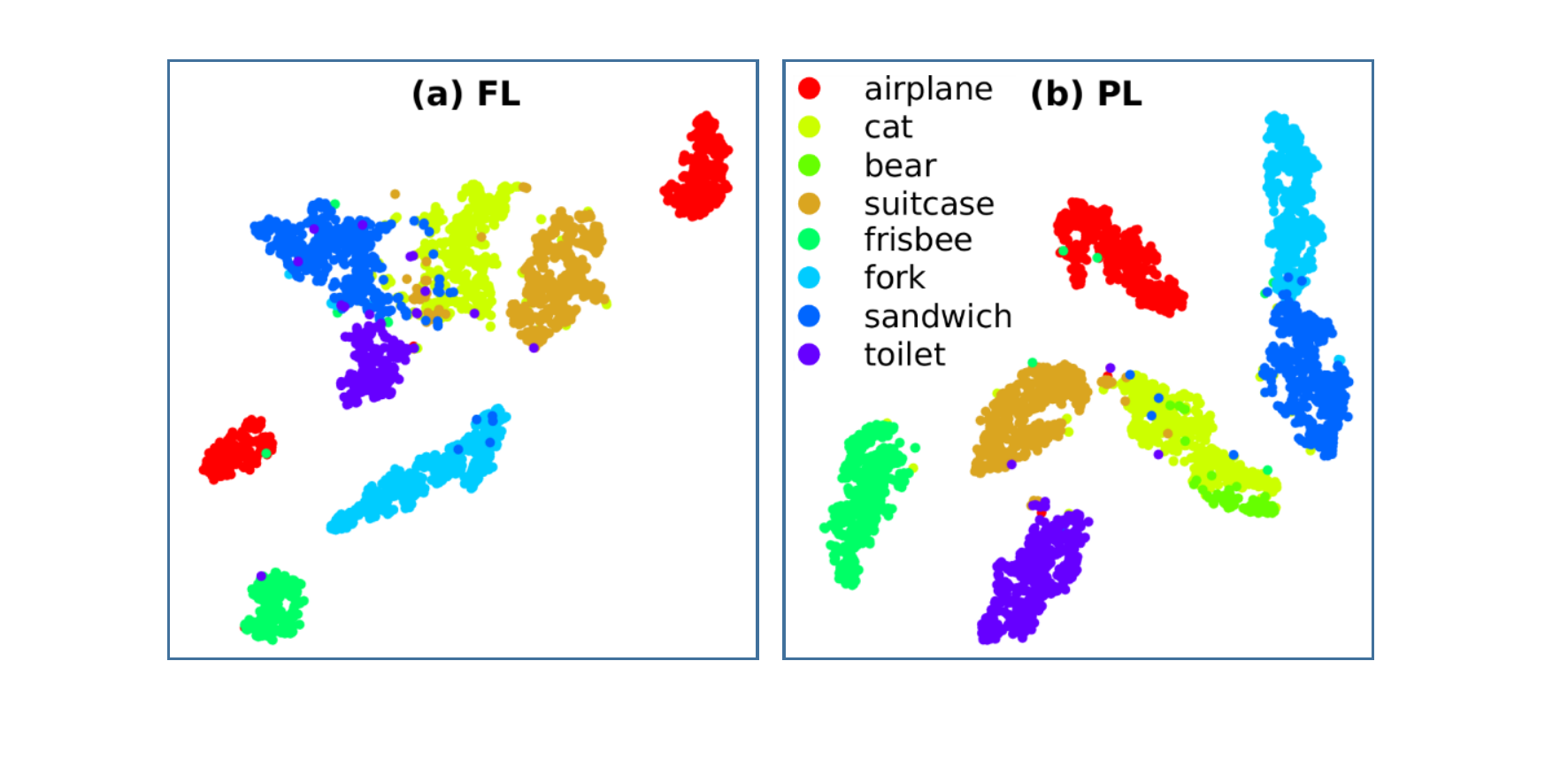}
  \caption{\small tSNE plot of visual features from 8 unseen classes projected onto semantic space using (a) FL \& (b) PL. FL pushes visual features close to their ground-truth. Thus, intra-class distances are minimized but inter-class distances are not considered. This works well 
  for seen class separation, but is not optimal for unseen classes because inter-class distances must be increased to ensure unseen class separability.
  Our PL ensures this requisite. 
  }
\label{fig:alignment_vis}
\end{figure}

 \textbf{Polarity Loss Properties:} The PL has two intriguing properties. 
\textit{(a) Word-vectors alignment:} For ZSL, generally visual features are projected onto the semantic word vectors. A high projection score indicates proper alignment with a word-vector. The overall goal of training is to achieve good alignment between the visual feature and its corresponding word-vector and an inverse alignment with all other word-vectors. In our proposed loss, $\text{FL}(\cdot)$ and $f_p$ perform the direct and inverse alignment respectively. Fig. \ref{fig:alignment_vis} shows visual features before and after this alignment. \textit{(b) Class imbalance:} The penalty function $f_p$ follows a trend similar to $\alpha_t$ and $(1{-}p_t)^\gamma$. It means that $f_p$ assigns a low penalty to well-classified/easy examples and a high penalty to poorly-performed/hard cases. It greatly helps in tackling class imbalance for single stage detectors where negative boxes heavily outnumber positive detections.

\begin{figure}[!t]
  \centering
   \includegraphics[width=1\columnwidth,trim={1cm 0cm 1cm .3cm},clip]{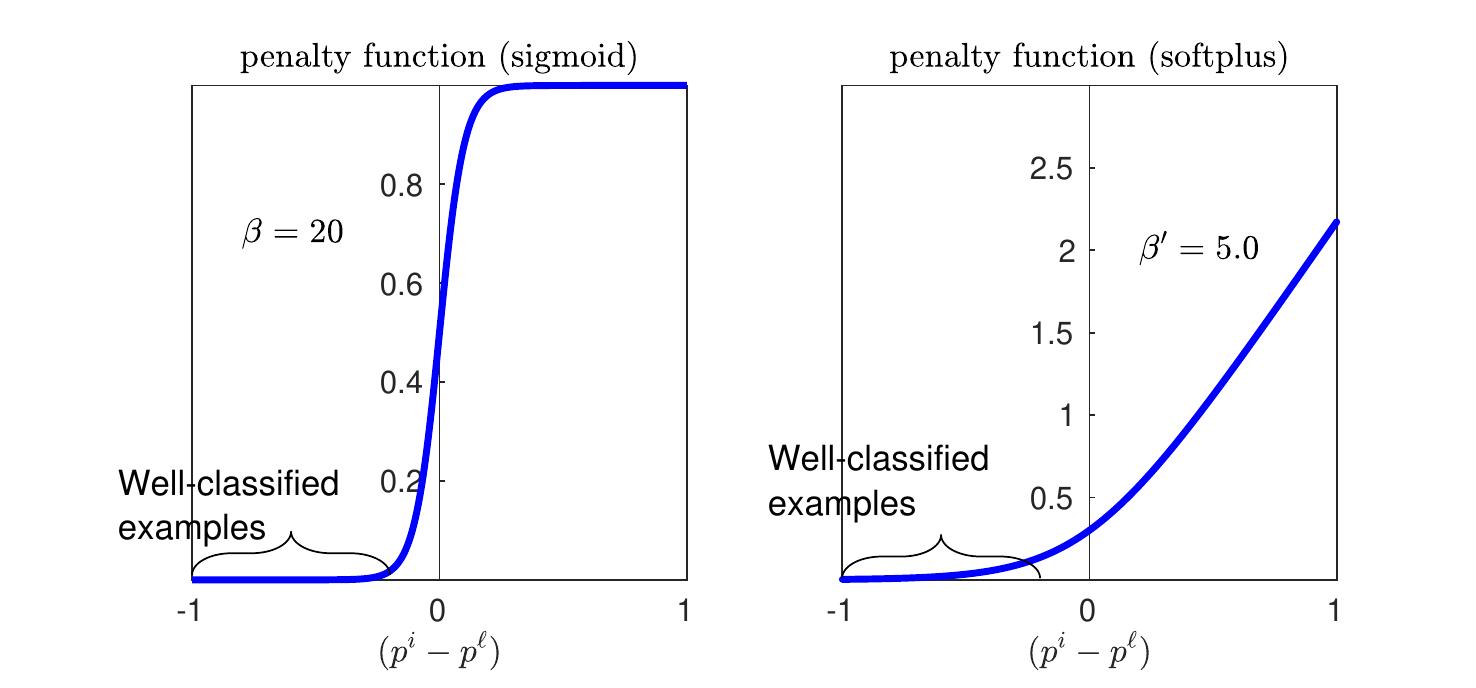}
   \caption{\small Visualization of sigmoid (left) vs. softplus (right) based penalty functions.}
\label{fig:losses2}
\end{figure}

 \textbf{Alternative formulation of polarity loss:} \label{sec:alternative}
We have used a sigmoid based penalty function to implement $f_p(p^i-p^{\ell})$ in our proposed polarity loss. Now, we present an alternative implementation of the penalty function based on the softplus function. In Fig.~\ref{fig:losses2}, we illustrate the shapes of both sigmoid and softplus based penalty functions. Both the functions increase penalty when $p^i-p^{\ell}$ moves from $-1$ to $1$. However, the softplus is relatively smoother than sigmoid. Also, softplus has a flexibility to assign a penalty $>1$ for any poorly classified examples whereas sigmoid can penalize at most 1. The formulation for the softplus based penalty function is as follows,
\begin{align}
f_p(p^i-p^{\ell}) = \log \Big( 1+\text{e}^{\beta' (p^i-p^{\ell})} \Big)
\label{eq:penalty2}
\end{align}
where, $\beta'$ is the loss hyper-parameter. The final polarity loss with the softplus based penalty function is the following:
\begin{align}
\mathscr{L}_{PL}&(\mathbf{p},\mathbf{y}) = \sum_i -\alpha^i_t (1-p^i_t)^\gamma \log \Big( 1+\text{e}^{\beta' (p^i-p^{\ell})} \Big) \log p^i_t, \text{  } \notag\\
& p^i_t =\begin{cases}
    p^i, & \text{if $y^i=1$}\\
    1-p^i, & \text{otherwise}
  \end{cases}\quad
    p^{\ell} = p^{i} \llbracket y^{i}=1 \rrbracket,
\end{align}


\subsection{Vocabulary Metric Learning}\label{sec:vocab_met}
Apart from proper visual-semantic alignment and class imbalance, a significant challenge for ZSD is the inherent noise in the semantic space.  In this paper, we propose a new `vocabulary metric learning' approach to improve the quality of word vectors for ZSL tasks. For brevity of expression, we restrict our discussion to the case of classification probability prediction or bounding box regression for a single anchor. For the classification case, 
metric learning is considered as a part of Polarity loss in Eq.~\ref{eq:final_obj}. Suppose, the visual feature of that anchor, $\mathbf{a}$ is $\phi(\mathbf{a}) = \mathbf{f}$, where $\phi$ represents the detector network.  The total number of seen classes is $\mathrm{S}$  and a matrix $W_s \in \mathbb{R}^{\mathrm{S}\times d}$ denotes all the $d$-dimensional word vectors of $\mathrm{S}$ seen classes arranged row-wise. The detector network $\phi$ is augmented with FC layers towards the head to transform the visual feature $\mathbf{f}$ to have the same dimension as the 
word vectors, i.e., $\mathbf{f} \in \mathbb{R}^{d}$. In Fig.~\ref{fig:arch}, we describe several ways to learn the alignment function between visual features and semantic information. We elaborate these further below.

\subsubsection{Learning with Word-vectors}
\begin{figure}[t]
  \centering   \includegraphics[width=1\columnwidth,trim={0cm 0cm 0cm 0cm},clip]{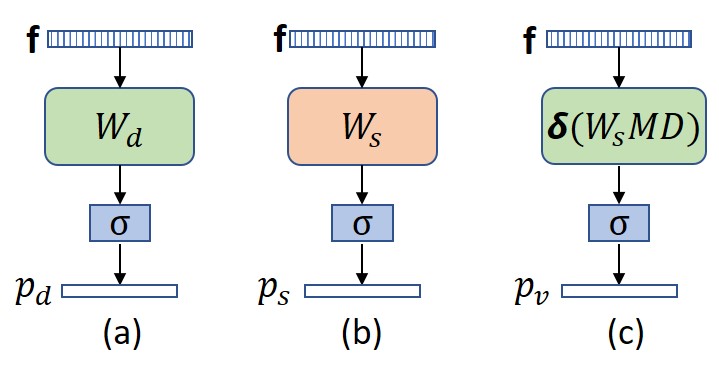}
  \caption{\small (a) Traditional basic approach with learnable $W_d$, (b) Inserting word vectors as a fixed embedding $W_s$, (c) learnable word vectors with vocabulary metric $\delta(W_s M D)$. }
\label{fig:arch}
\end{figure}

For the traditional detection case, shown in Fig. \ref{fig:arch}(a), the visual features $\mathbf{f}$ are transformed with a learnable FC layer $W_d \in \mathbb{R}^{\mathrm{S}\times d}$, followed by a sigmoid/softmax activation ($\sigma$) to calculate $\mathrm{S}$ prediction probabilities, $\mathbf{p}_d = \sigma(W_d \mathbf{f})$. This approach works well for traditional object detection, but it is not suitable for the zero-shot setting as the transformation $W_d$ cannot work with unseen object classes. 

A simple extension of the traditional detection framework to the zero-shot setting is possible by replacing trainable weights of the FC layers, $W_d$, by the non-trainable seen word vectors $W_s$ (Fig. \ref{fig:arch}(b)). Keeping this layer frozen, we allow projection of the visual feature $\mathbf{f}$ to the word embedding space to calculate prediction scores $\mathbf{p}_s$: 
\begin{align}
\mathbf{p}_s = \sigma(W_s \mathbf{f})
\label{eq:withoutvocab}
\end{align}
This projection aligns visual features with the word vector of the corresponding true class. The intuition is that rather than directly learning a prediction score from visual features (in Fig \ref{fig:arch}(a)), it is better to learn a correspondence between the visual features with word vectors before the prediction.

\textbf{Challenges with Basic Approach: } Although the configuration described in Fig.~\ref{fig:arch}(b) delivers a basic solution to zero-shot detection, it suffers from several limitations. (1) \textit{Fixed Representations:} With a fixed embedding $W_s$, the network cannot update the semantic representations and has limited flexibility to properly align visual and semantic domains. (2) \textit{Limited word embeddings:} The word embedding space is usually learned using billions of words from unannotated texts which results in noisy word embeddings. Understanding the semantic space with only $\mathrm{S}$ word vectors is therefore unstable and insufficient to model visual-semantic relationships. (3) \textit{Unseen-background confusion:} In ZSD, one common problem is that the model confuses unseen objects with background since it has not seen any visual instances of unseen classes \cite{Bansal_2018_ECCV}.

\begin{figure}[t]
  \centering
  \includegraphics[width=1\columnwidth,trim={2.2cm .05cm 1.3cm .8cm},clip]{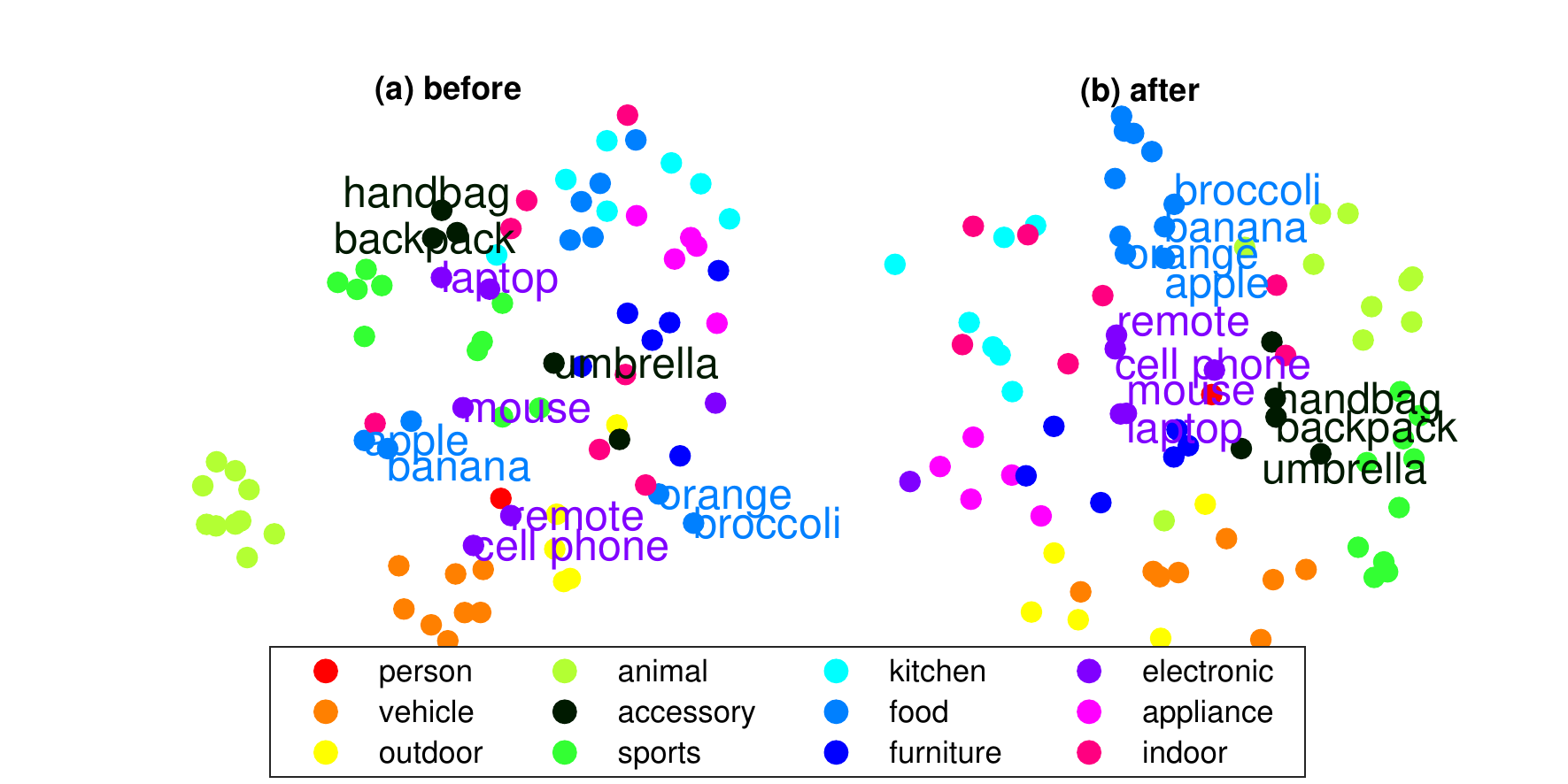}
  \caption{\small 2D tSNE~\cite{tSNE_van2014} embedding of word2vec: (a) before (b) after modification based on vocabulary metric with our loss. Word-vectors are more evenly distributed in (b) than (a). Also, visually similar classes for example, apple/banana/orange/broccoli, cell phone/remote/laptop/mouse and handbag/backpack/umbrella are embedded more closely in (b) than (a). Super-category annotations are used for visualization only, not during our training.}
\label{fig:modified_word}
\end{figure}

\begin{figure*}[!htp]
  \centering
  \includegraphics[trim=0 0 0 3mm, clip, width=1\textwidth]{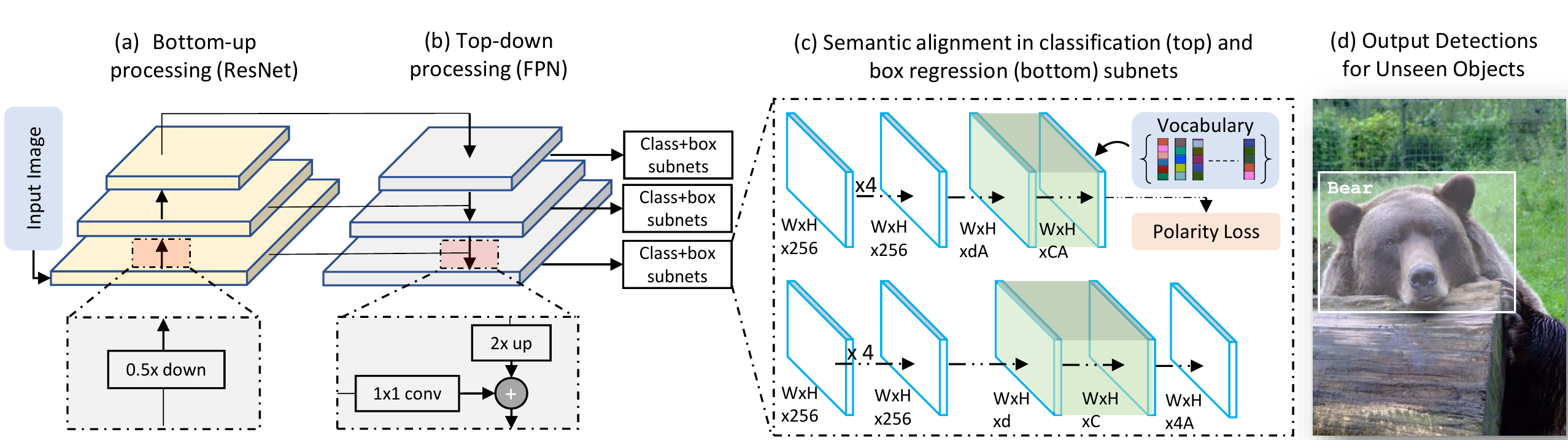}

  \caption{\small Network architecture for ZSD. The green colored layer implements Eq. \ref{eq:withoutvocab} (\method{Our-PL-word}) or \ref{eq:vocab} (\method{Our-PL-vocab}).} 
   
\label{fig:arch2}
\end{figure*}

\subsubsection{Learning with vocabulary metric}

To address the above gaps, we propose to learn a more expressive and flexible semantic domain representation. Such a representation can lead to a better alignment between visual features and word vectors. 
Precisely, we propose a vocabulary metric method summarized in Fig.~\ref{fig:arch}(c) that takes advantage of the word vectors of a pre-defined vocabulary, $D \in \mathbb{R}^{v \times d}$ (where $v$ is the number of words in the vocabulary). By relating the given class semantics with dictionary atoms, the proposed approach  provides 
an inherent mechanism to update class-semantics optimally for ZSD. Now, we calculate the prediction score as follows:
\begin{align}
	\mathbf{p}_v = \sigma( \delta(W_s M D) \mathbf{f})
    \label{eq:vocab}
\end{align}
Here, $M \in \mathbb{R}^{d \times v}$ represents the learnable parameters which connect seen word vectors with the vocabulary and $\delta(.)$ is a $\tanh$ activation function. $M$ can be interpreted as learned attention over the dictionary.  With such an attention, the network can understand the semantic space better and learn a rich representation because it considers more linguistic examples (vocabulary words) inside the semantic space.
Simultaneously, it helps the network to update the word embedding space for better alignment with visual features. 
Further, it reduces unseen-background confusion since the network can relate visual features more accurately with a diverse set of linguistic concepts. We visualize word vectors before and after the update in Fig.~\ref{fig:modified_word} (a) and (b) respectively.

Here, we emphasize that the previous attempts to use such an external vocabulary have their respective limitations. For example, \cite{Al-Halah_2016_CVPR} considered a limited set of attributes while \cite{al2017automatic} used several disjoint training stages. These approaches are therefore not end-to-end trainable. Further, they only investigate the recognition problem.

\textbf{Regression branch with semantics:} Eq. \ref{eq:vocab} allows our network to predict seen class probabilities at the classification branch directly using semantic information from the vocabulary metric. Similarly, we also apply such semantics in the regression branch with some additional trainable FC layers. In our experiments, we show that adding semantics in this manner leads to further improvement in ZSD. It shows that the predicted regression box can benefit from the semantic information that improves the overall performance.

\section{Architecture details} \label{sec:architecture}
\textbf{Single-stage Detector:} Our proposed ZSD framework is specially designed to work with single-stage detectors. The primary motivation is the direct connection between anchor classification and localization that ensures a strong feedback for both tasks. For this study, we choose a recent unified single architecture, RetinaNet~\cite{RetinaNet_2017_ICCV} to implement our proposed method. RetinaNet is the best detector known for its high speed (on par with single-stage detectors) and high accuracy (outperforming two-stage detectors). In Fig. \ref{fig:arch2}, we illustrate the overall architecture of the model. In addition to a novel loss formulation, we also perform modifications to the RetinaNet architecture to link visual features (from ResNet50~\cite{ResNet_CVPR_2016}) with semantic information. 
To adapt this network to ZSL setting, we perform simple modifications in both classification and box regression subnets to consider word-vectors (with vocabulary metric) during training (see Fig.~\ref{fig:arch2}). 

RetinaNet has one backbone network called Feature Pyramid Network (FPN)~\cite{FPN_2017_CVPR} and two task-specific subnetwork branches for classification and box regression. FPN extracts rich, multi-scale features for different anchor boxes from an image to detect objects at different scales. For each pyramid level, we use anchors at \{1:2,1:1,2:1\} aspect ratios with sizes $\{2^0,2^{1/3},2^{2/3}\}$ totaling to $A{=}9$ anchors per level, covering an area of $32^2$ to $512^2$ pixels. The classification and box-regression subnetworks attempt to predict the one-hot target ground-truth vector of size $\mathrm{S}$ and box parameters of size four respectively. We consider an anchor box as an object if it gets an intersection-over-union (IoU) ratio $>0.5$ with a ground-truth bounding box. 

 \textbf{Modifications to RetinaNet:}
Suppose, a feature map at a given pyramid level has $C$ channels. For the \textbf{classification subnet}, we first apply four conv layers with $C$  filters of size $3\times 3$, followed by ReLU, similar to RetinaNet. Afterward, we apply a $3 \times 3$ conv layer with $d\times A$ filters to convert visual features to the dimension of word vectors, $d$. Next, we apply a custom layer which projects image features onto the word vectors. We also apply a sigmoid activation function to the output of the projection. This custom layer may have fixed parameters like Fig.~3(b) or trainable parameters like 3(c) of the main paper with vocabulary metric. These operations are formulated as: $\mathbf{p}_s = \sigma(W_s \mathbf{f})$ or $\mathbf{p}_v = \sigma( \delta(W_s M D) \mathbf{f})$ depending on the implementation of the custom layer. Similarly, for the \textbf{box-regression} branch, we attach another $3 \times 3$ convolution layer with $C$ filters and ReLU non-linearity, followed by $3 \times 3$ convolution with $d$ filters and the custom layer to get the projection response. Finally, another convolution with $4A$ to predict a relative offset between the anchor and ground-truth box. In this way, the box-prediction branch gets semantic information of word-vectors to predict offsets for regression. Note that, similar to \cite{RetinaNet_2017_ICCV}, the classification and regression branches do not share any parameters, however, they have a similar structure.

 \textbf{Training:} We train the classification subnet branch with our proposed loss defined in Eq.~\ref{eq:final_obj}. Similar to \cite{RetinaNet_2017_ICCV}, to address the imbalance between hard and easy examples, we normalize the total classification loss (calculated from $\sim$100k anchors) by the total number of object/positive anchor boxes rather than the total number of anchors. We use standard smooth $L_1$ loss for the box-regression subnet branch. The total loss is the sum of the loss of both branches. 
 
 \begin{figure*}[!ht]
  \centering
   \includegraphics[width=1\textwidth,trim={5cm 0cm .3cm 0cm},clip]{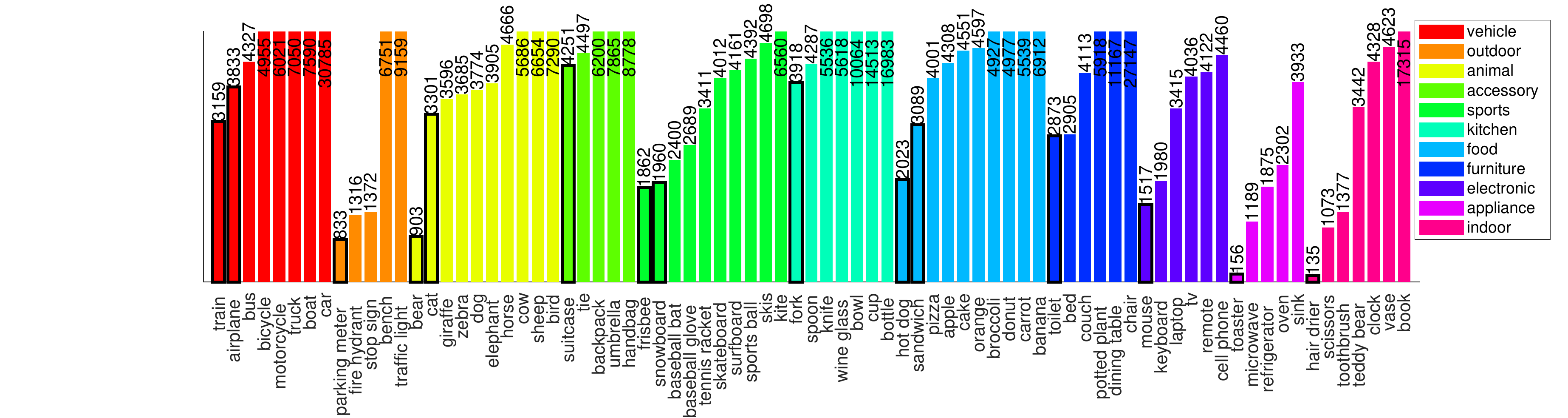}
   \caption{\small Instances of each class in the MS-COCO dataset (except class `person'). Tall bars are clipped for better visualization. The class bars with black border are selected as unseen classes. We choose 20\% rarest classes from each superclass as unseen.}
\label{fig:cocostat}
\end{figure*}

 \textbf{Inference:} For seen object detection, a simple forward pass predicts both confidence scores and bounding boxes in the classification and box-regression subnetworks respectively. Note that we only consider a fixed number (e.g., 100) of boxes from RPN having confidence greater than 0.05 for inference.
Moreover, we apply Non-Maximum Suppression (NMS) with a threshold of 0.5 to obtain final detections. We select the final detections that satisfy a seen score-threshold ($t_s$). To detect unseen objects, we use the following equation, followed by an unseen score-thresholding with a relatively lower value ($t_u < t_s$)\footnote{Empirically, we found $t_s{=}0.3$ and $t_u{=}0.1$ generally work well. $t_u$ is kept smaller to counter classifier bias towards unseen classes due to the lack of visual examples during training. }:
\begin{align}
	\mathbf{p}_u = W_u W_s^T\sigma( \delta(W_s M D) \mathbf{f})
    \label{eq:unseenpred}
\end{align} 
where, $W_u \in \mathbb{R}^{\mathrm{U}\times d}$ contains unseen class word vectors. For generalized zero-shot object detection (GZSD), we simply consider all detected seen and unseen objects together. In our experiments, we report performances for traditional seen, zero-shot unseen detection and GZSD. One can notice that our architecture predicts a bounding box for every anchor which is independent of seen classes. It enables the network to predict bounding boxes dedicated to unseen objects. Previous attempts like \cite{rahman2018ZSD} detect seen objects first and then attempt to classify those detections to unseen objects based on semantic similarity. By contrast, our model allows detection of unseen bounding boxes that are different to those seen.

 \textbf{Reduced description of unseen:} All seen semantics vectors are not necessary to describe an unseen objects \cite{Rahman_TIP_2018}. Thus, we only consider the top $T$ predictions, $\mathbf{p}_v' \in \mathbb{R}^{T}$ from $\sigma( \delta(W_s M D) \mathbf{f})$ and the corresponding seen word vectors, $W_s' \in \mathbb{R}^{\mathrm{T}\times d}$ to predict unseen scores. For the reduced case,
$
    \mathbf{p}_u' = W_u W_s'^T \mathbf{p}_v'.
$
In the experiments, we vary the number of the closest seen $T$ from 5 to $\mathrm{S}$ and find that a relatively small value of $T$ (e.g., 5) performs better than using all available $T = \mathrm{S}$ seen word vectors.

\begin{figure*}[!htp]
\begin{minipage}{.68\textwidth}
\scalebox{.95}{
\begin{tabular}{|c|>{\columncolor{clr4}}c|c|c|c|c|}
\hline
\rowcolor{clr3}  \multicolumn{1}{|l|}{\textbf{Method}}  & {\textbf{Seen /}}  &  & \multicolumn{3}{c|}{\textbf{GZSD}}\\ \cline{4-6} 
\rowcolor{clr3}  & {\textbf{Unseen}}&  {\textbf{ZSD}}  & Seen	& Unseen & HM \\
\rowcolor{clr3} \multicolumn{1}{|l|}{Split in \cite{Bansal_2018_ECCV} ($\downarrow$)} &   & (mAP/RE) & (mAP/RE)	& (mAP/RE) & (mAP/RE) \\ \hline
SB~\cite{Bansal_2018_ECCV} &48/17&0.70/24.39&-&-&- \\ \hline
DSES~\cite{Bansal_2018_ECCV} &48/17&0.54/27.19&-/15.02&-/15.32&-/15.17 \\ \hline
ZSD-Textual~\cite{Li_AAAI_2019}&48/17&-/34.3&-/-&-/-&-/-\\\hline
\method{Baseline}&48/17 	&6.99/18.65&40.46/43.69&2.88/17.89&5.38/25.38 \\ \hline
\method{Ours}&48/17 & \textbf{10.01/43.56}&35.92/38.24&\textbf{4.12/26.32}&\textbf{7.39/31.18} \\ \hline
\hline
\rowcolor{clr3} \multicolumn{2}{|l|}{Proposed Split ($\downarrow$)} & mAP/RE & mAP/RE & mAP/RE & mAP/RE \\
\hline 
\method{Baseline}&65/15 & 8.48/20.44&\textbf{36.96/40.09}&8.66/20.45&14.03/27.08  \\ \hline
\method{Ours}&65/15 &\textbf{12.40/37.72} & 34.07/36.38 & \textbf{12.40/37.16} & \textbf{18.18/36.76} \\ \hline
\end{tabular}}
\end{minipage}\hfill
\begin{minipage}{.32\textwidth}
\includegraphics[width=1\textwidth,trim={0cm 0cm 0cm 0cm},clip]{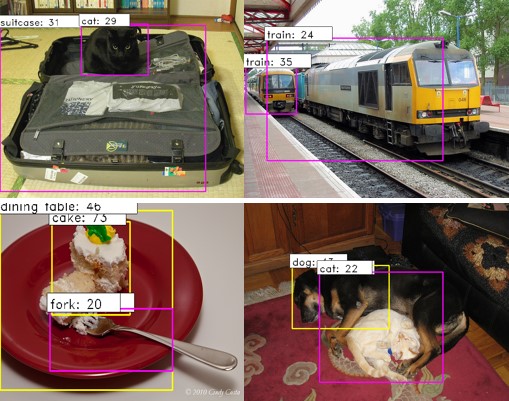}
\end{minipage}
{
\caption{\small \textit{(left)} Overall performance on MS-COCO. Hyper-parameters are set on the validation set: $\beta{=}5$, IoU${=} 0.5$. mAP = mean average precision and RE = recall (@100). The top part shows results on 
\cite{Bansal_2018_ECCV} split and the lower part shows results on our proposed split. \method{Ours} achieves best performance in terms of mAP on unseen classes. \textit{(Right)} Qualitative examples of ZSD (top row) and GZSD (bottom row). Pink and yellow box represent unseen and seen detections respectively. More qualitative results are presented in Fig.~\ref{fig:zsd2}.}
\label{tab:overall}
}
\end{figure*}

\section{Experiments}\label{sec:exp}

    \textbf{Datasets:} We evaluate our method with 
    MS-COCO($2014$)~\cite{MSCOCO_2014} and 
    Pascal VOC ($2007/12$)~\cite{VOC_IJCV_2010}. With $80$ object classes, MS-COCO includes \num{82783} training and \num{40504} validation images. For the ZSD task, only unseen class performance is of interest. As the test data labels are not known, the ZSD evaluation is done on a subset of validation data. MS-COCO($2014$) has more validation images than any later versions which motivates us to use it. For Pascal VOC, we use the train set of $2007$ and $2012$ for training and use validation+test set of $2007$ for testing.

  \textbf{Issues with existing MS-COCO split:} Recently, \cite{Bansal_2018_ECCV} proposed a split of seen/unseen classes for MS-COCO (2014). It considers $73,774$ training images from $48$ seen classes and \num{6608} test images from $17$ unseen classes. The split criteria were the cluster embedding of class semantics and synset WordNet hierarchy~\cite{Wordnet_1995}. We identify two practical {drawbacks} of this split: \textbf{(1)} Because all $63$ classes are not used as seen, this split does not take full advantage of training images/annotations, \textbf{(2)} Because of choosing unseen classes based on wordvector clustering it cannot guarantee the desired diverse nature of the unseen set. For example, this split does not choose any classes from `\emph{outdoor}' super-category of MS-COCO.

  \textbf{Proposed seen/unseen split on MS-COCO:} To address these issues, we propose a more realistic split of MS-COCO for ZSD. Following the practical consideration of unseen classes discussed in \cite{rahman2018ZSD} i.e. rarity and diverseness, we follow the following steps: \textbf{(1)} We sort classes of each superclass in ascending order based on the total number of instances in the training set. \textbf{(2)} For each superclass, 
 we pick $20\%$ rare classes as unseen which results in $15$ unseen and $65$ seen classes. Note that the superclass information is only used to create a diverse seen/unseen split, and never used during training. \textbf{(3)} Being zero-shot, we remove all the images from the training set where at least one unseen class appears to create a training set of \num{62300} images. \textbf{(4)} For testing {ZSD}, we select \num{10098} images from the validation set where at least one instance of an unseen class is present. The total number of unseen bounding boxes is \num{16388}. {We use both seen and unseen annotation together for this set to perform GZSD}. \textbf{(5)} We prepare another list of \num{38096} images from the validation set where at least one occurrence of the seen instance is present to test traditional detection performance on seen classes.
 
 In Fig.~\ref{fig:cocostat}, we present all 80 MS-COCO classes in sorted order across each super-category based on the number of instance/bounding boxes inside the training set. Choosing 20\% low-instance classes from each super-category ensures the rarity and diverseness for the chosen unseen classes. In this paper, we report results on both our and \cite{Bansal_2018_ECCV} settings.

\begin{table*}[!t]
  \centering\setlength\tabcolsep{2.5pt}
  \scalebox{1}{
    \begin{tabular}{|c|c|c||c|c|c|c|c|c|c|c|c|c|c|c|c|c|c|c||c|c|c|c|}
    \hline
\rowcolor{clr3}    
\rotatebox{90}{\textbf{Method}}&\rotatebox{90}{{ \textbf{Seen}}}&\rotatebox{90}{{ \textbf{Unseen}}}&\rotatebox{90}{aeroplane}&\rotatebox{90}{bicycle}&\rotatebox{90}{bird}&\rotatebox{90}{boat}&\rotatebox{90}{bottle}&\rotatebox{90}{bus}&\rotatebox{90}{cat}&\rotatebox{90}{chair}&\rotatebox{90}{cow}&\rotatebox{90}{d.table}&\rotatebox{90}{horse}&\rotatebox{90}{motrobike}&\rotatebox{90}{person}&\rotatebox{90}{p.plant}&\rotatebox{90}{sheep}&\rotatebox{90}{tvmonitor}&\rotatebox{90}{{\underline{\textit{car}}}}&\rotatebox{90}{\underline{\textit{dog}}}&\rotatebox{90}{\underline{\textit{sofa}}}&\rotatebox{90}{\underline{\textit{train}}} \\  \hline

Demirel \emph{et al.}\cite{Demirel_BMVC_2018}&57.9&54.5&68.0&\textbf{72.0}&\textbf{74.0}&48.0&41.0&61.0&48.0&25.0&48.0&\textbf{73.0}&\textbf{75.0}&71.0&73.0&33.0&59.0&57.0&55.0&82.0&\textbf{55.0}&26.0 \\ \hline
\method{Ours} &\textbf{63.5}&\textbf{62.1}&\textbf{74.4}&71.2&67.0&\textbf{50.1}&\textbf{50.8}&\textbf{67.6}&\textbf{84.7}&\textbf{44.8}&\textbf{68.6}&39.6&74.9&\textbf{76.0}&\textbf{79.5}&\textbf{39.6}&\textbf{61.6}&\textbf{66.1}&\textbf{63.7}&\textbf{87.2}&53.2&\textbf{44.1} \\ \hline

    \end{tabular}}
  \caption{\small mAP scores of Pascal VOC'07. \underline{\textit{Italic}} classes are unseen.} 
  \label{tab:Demirel}
\end{table*}

\textbf{Pascal VOC Split:} For Pascal VOC 2007/12~\cite{VOC_IJCV_2010}, we follow the settings of \cite{Demirel_BMVC_2018}. We use $16$ seen and $4$ unseen classes from total $20$ classes. We utilize \num{2072} and \num{3909} train images from Pascal VOC $2007$ and $2012$ respectively after ignoring images containing any instance of unseen classes. For testing, we use \num{1402} val+test images from Pascal VOC 2007 where any unseen class appears at least once.

\textbf{Vocabulary:} We choose vocabulary atoms from \num{5018} Flickr tags in NUS-WIDE~\cite{NUS_WIDE_09}. We only remove MS-COCO class names and tags that have no word vectors. This vocabulary covers a wide variety of objects, attributes, scene types, actions, and visual concepts.
   
\textbf{Semantic embedding:} For MS-COCO classes and vocabulary words, we use $\ell_2$ normalized $300$ dimensional unsupervised word2vec \cite{Mikolov_NIPS_2013}, GloVe \cite{Jeffrey_Glove_2014} and FastText \cite{fasttext_2017} vectors obtained from billions of words from unannotated texts like Wikipedia. For Pascal VOC~\cite{VOC_IJCV_2010} classes, we use average 64 dimension binary per-instance attribute annotation of all training images from aPY dataset \cite{aPY_2009}. Unless mentioned otherwise, we use word2vec in our experiments.

\textbf{Evaluation metric:} Being an object detection problem, we evaluate using mean average precision (mAP) at a particular IoU. Unless mentioned otherwise, we use IoU$= 0.5$. 
Notably, \cite{Demirel_BMVC_2018} use the Recall measure for evaluations, however since recall based evaluation does not penalize a method for the wrongly predicted bounding boxes, we only recommend mAP based evaluation for ZSD. To evaluate GZSD, we report the harmonic mean (HM) of mAP and recall \cite{Bansal_2018_ECCV,Xian_PAMI_2018}.

{\textbf{Implementation details:}} We implement FPN with a basic ResNet50~\cite{ResNet_CVPR_2016}. All images are rescaled to make their smallest side $800$px. We train the \texttt{FL-basic} method using the original RetinaNet architecture with only training images of seen classes so that the pre-trained network does not get influenced by unseen instances. Then, to train \texttt{Our-FL-word}, we use the pre-trained weights to initialize the common layers of our framework. We initialize all other uncommon layers with a uniform random distribution. Similarly, we train \texttt{Our-PL-word} and \texttt{Our-FL-vocab} upon the training of \texttt{Our-FL-word}. Finally, we train \texttt{Our-PL-vocab} using the pre-trained network of \texttt{Our-FL-vocab}. We train each network for $500$k iterations keeping a single image in the minibatch. The only exception while training with our proposed loss is to train for $100$k iterations instead of $500$k. Each training time varies from $72$ to $96$ hours using a single Tesla P$100$ GPU. For optimization, we use Adam optimizer with learning rate $10^{-5}$, $\beta_1=0.9$ and $\beta_2=0.999$. We implement this framework with \textit{Keras} library.

\begin{table}[!t]
  \centering
  \scalebox{1}{
    \begin{tabular}{|c|c|c|c|c|c|c|}
    \hline
    $\beta$($\rightarrow$) & 5  & 10 & 15 & 20 & 25 & 30 \\
    \hline
    mAP &\textbf{48.6}&47.9&47.8&47.6&47.5&47.9\\ \hline
    \end{tabular}}
  \caption{{Validation study on traditional detection.}}
  \label{tab:validation}
\end{table}

{\textbf{Validation strategy:} $\alpha, \gamma$ and $\beta$ are the hyper-parameters of the proposed polarity loss. Among them, $\alpha, \gamma$ are also present in the focal loss. Therefore, we choose $\alpha = 0.25, \gamma = 2.0$ as recommended by the original RetinaNet paper \cite{RetinaNet_2017_ICCV}. $\beta$  is the only new hyper-parameter introduced in the polarity loss. We conduct a validation experiment on seen classes to perform the traditional detection task. In Table \ref{tab:validation}, we tested $\beta = \{5,10,15,20,30\}$ and picked $\beta=5$ for our loss as it performed the best among all the considered values.}

\subsection{Quantitative Results}

{\textbf{Compared Methods:}}
We rigorously evaluate our proposed ZSD method on both 
\cite{Bansal_2018_ECCV} split ($48/17$) and our new ($65/15$) split of MS-COCO. We provide a brief description of all compared methods:
\textbf{(a)} SB~\cite{Bansal_2018_ECCV}: This method extracts pre-trained Inception-ResNet-v2 features from Edge-Box object proposals. It applies a standard max-margin loss to align visual features to semantic embeddings via linear projections. \textbf{(b)} DSES~\cite{Bansal_2018_ECCV}: In addition to SB, DSES augments extra bounding boxes other than MSCOCO objects. As \cite{Bansal_2018_ECCV} reported recall performances, we also report recall results (in addition to mAP) to compare with this method. \textbf{(c)} \method{Baseline}: This method trains an exact RetinaNet model. Thus, it does not use any word vectors during training. To extend this approach to perform ZSD, we apply this formula to calculate unseen scores: $\mathbf{p}_{u}'{=}W_u W_s'^T \mathbf{p}_{d}'$ where $\mathbf{p}_d'$ represents top T seen prediction scores for the reduced description of unseen.
\textbf{(d)} \method{Ours}: This method is our final proposal using vocabulary 
and polarity loss (Fig.~\ref{fig:arch2}).

{\textbf{Overall Results:}} Fig.~\ref{tab:overall} presents overall performance on ZSD and GZSD tasks across different comparison methods with two different seen/unseen split of MS-COCO. In addition to mAP, we also report recall (RE) to compare with \cite{Bansal_2018_ECCV}. With 48/17 settings, our method (and baseline) beats \cite{Bansal_2018_ECCV} (SB and DSES) in both the ZSD and GZSD by a significantly large margin. Similarly, in 65/15 split, we outperform our \method{baseline} by a margin 3.92 mAP (12.40 vs. 8.48) in ZSD task and 4.15 harmonic-mAP in GZSD task (18.18 vs. 14.03). This improvement is the result of end-to-end learning, the inclusion of the vocabulary metric to update word vectors and the proposed loss in our method. We report results on GloVe and FastText word vectors in the Table \ref{tab:diffword}

\begin{figure}
\begin{minipage}{.6\columnwidth}
    \scalebox{.8}{
    \begin{tabular}{|c|c|c|c|c|c|}
        \hline
        \multirow{2}{*}{$\mathbf{\gamma}$} &\multirow{2}{*}{$\mathbf{\alpha}$} &  & \multicolumn{3}{c|}{\textbf{GZSD}}\\ \cline{4-6}
          & &  {\textbf{ZSD}}  & Seen	& Unseen & HM \\ \hline
        0 & 1 &6.6&31.9&6.6&10.9\\
        0 & .75 &2.7&27.4&2.7&4.9\\
        0.1 & .75 &5.4&27.9&5.4&9.0\\
        0.2 & .75 &7.3&31.4&7.3&11.8\\
        0.5 & .50 &8.4&30.6&8.4&13.1\\
        1.0 & .25 &11.6&31.3&11.6&16.9\\
        2.0 & .25 &\textbf{12.6}&33.0&\textbf{12.6}&\textbf{18.3}\\
        5.0 & .25 &9.1&\textbf{33.6}&9.1&14.3\\ \hline
    \end{tabular}}
\end{minipage}\hfill
\begin{minipage}{.4\columnwidth}
    \includegraphics[width=1\columnwidth,trim={0cm 0cm 0cm 0cm},clip]{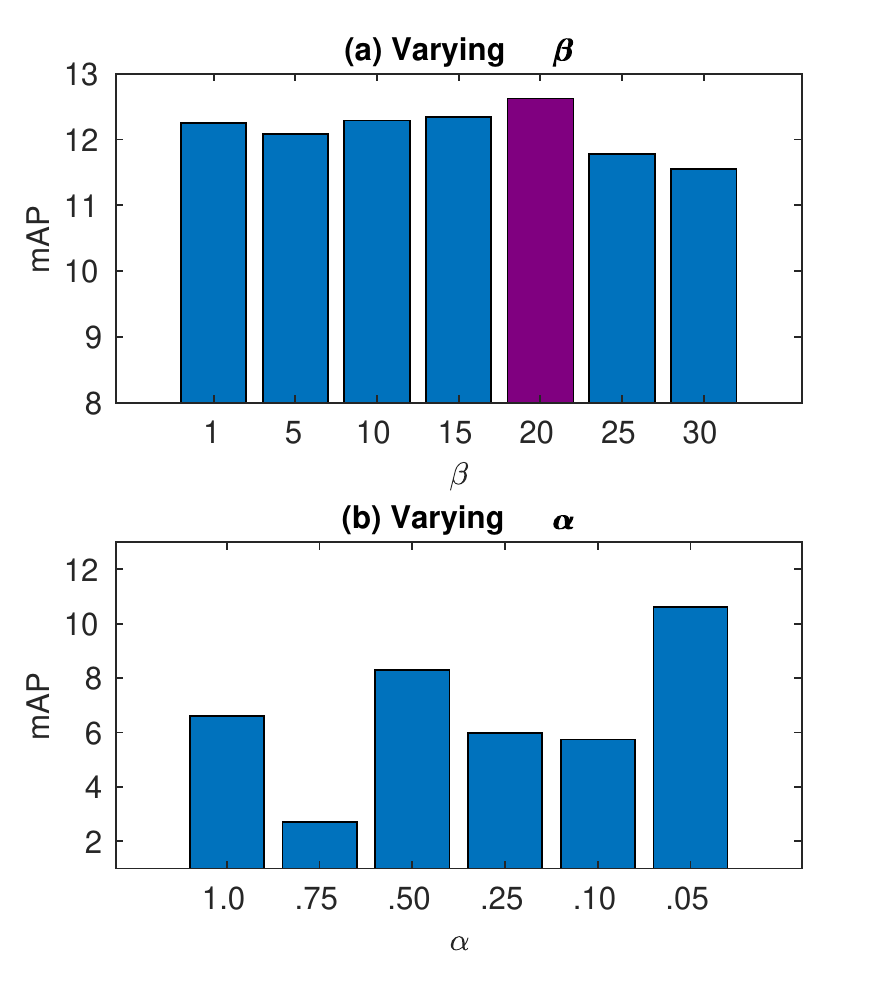}
\end{minipage}
{
 \caption{\small Parameter sensitivity analysis: \emph{(Left)} Varying $\alpha$ and $\gamma$ with a fixed $\beta{=}20$. \emph{(Right-a)} Impact of varying $\beta$, \emph{(Right-b)} varying $\alpha$ with $\gamma{=}0$ to see the behavior of our loss with only balanced CE. Note that the actual hyper-parameters choice is made on a validation set.}
 \label{fig:sensitivity}
}
\end{figure}

{\textbf{Hyper-parameter Sensitivity Analysis}:} We study the sensitivity of our model to loss hyper-parameters $\gamma, \alpha$ and $\beta$. First, we vary $\gamma \in [0,5]$  and $\alpha \in [.25,1]$ keeping $\beta{=}20$. In Fig.~\ref{fig:sensitivity} (\emph{left}), we report mAP 
using different parameter settings for ZSD and GZSD. Our model works best with $\alpha {=} .25$ and $\gamma {=} 2.0$ which are also the recommended values in FL. We also vary $\beta$ from $1$-$30$ to see its effect on ZSD in Fig.~\ref{fig:sensitivity} (\emph{Right-a}). This parameter controls the steepness of the penalty function $f_p$ in Eq. \ref{eq:penalty}. 
Notably $\beta {=}20$ provides correct steepness to estimate a penalty for incorrect predictions. Our loss can also work reasonably well with balanced CE (i.e., without FL when $\gamma{=}0$). We show this in Fig.~\ref{fig:sensitivity}(\emph{Right-b}). With a low $\alpha$ of $0.05$, our method can achieve around $10\%$ mAP. It shows that our penalty function can effectively balance object/background and easy/hard cases. 

\begin{figure}
    \begin{minipage}{.72\columnwidth}
        \centering
        \scalebox{.75}{
        \begin{tabular}{|c|c|c|c|c|}
        \hline
         \multirow{2}{*} {\textbf{Method}} & \multirow{2}{*}{\textbf{ZSD}} & \multicolumn{3}{c|}{\textbf{GZSD}}\\ \cline{3-5} 
         &    & Seen	& Unseen & HM \\
        
        \hline 
        \method{Baseline} & 8.48&\textbf{36.96}&8.66&14.03  \\ \hline
        \method{Our-FL-word}  & 10.80& 37.56& 10.80& 16.77 \\ \hline
        \method{{Our-}PL-word} &	12.02&33.28&12.02&17.66\\ \hline
        \method{{Our-}PL-vocab}*&\textbf{12.62} & 32.99	& \textbf{12.62} & \textbf{18.26} \\ \hline
        \end{tabular}}
    \end{minipage}
    \hfill
    \begin{minipage}{0.25\columnwidth}
        \centering
        \includegraphics[width=\columnwidth,trim={3.8cm 0cm 4.3cm 0cm},clip]{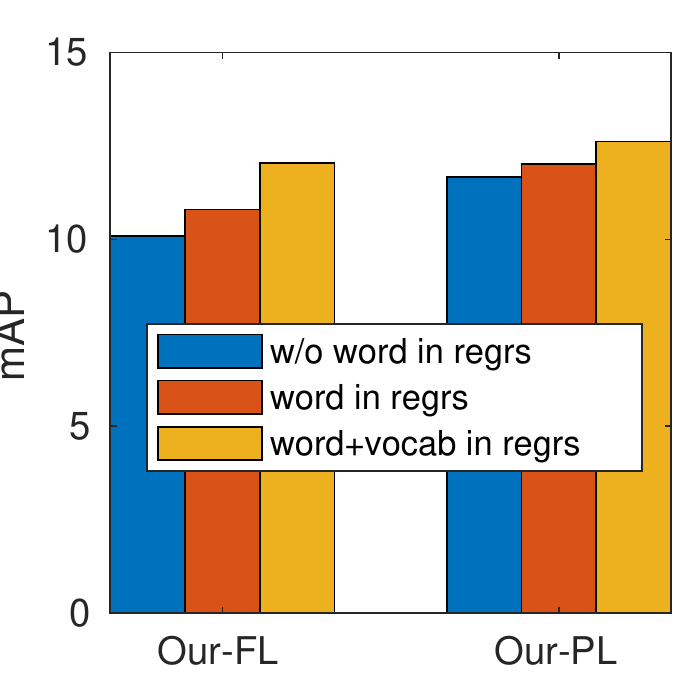}
    \end{minipage}
{
 \caption{\small Ablation studies with $\beta=20$: \emph{(Left)} Comparison of different variant of our approach, best method denoted with $*$. \emph{(Right)} Impact of word-vectors in the regression branch.}
 \label{fig:ablation}
}
\end{figure}

{\textbf{Ablation Studies:}} In Fig. \ref{fig:ablation}\textit{(Left)}, we report results on different variants of our method. \method{Our-FL-word}: This method is based on the architecture in Fig. \ref{fig:arch}(b) and trained with focal loss. It uses static word vectors during training. But, it cannot update vectors based on visual features. \method{Our-PL-word}: Same architecture as of \method{Our-FL-word} but training is done with our proposed polarity loss. \method{Our-PL-vocab}: The method uses our proposed framework in Fig.~\ref{fig:arch2} with vocabulary metric learning in the custom layer and is learned with polarity loss (without vocabulary metric learning). Our observations: \textbf{(1)} \method{Our-FL-word} works better than \method{Baseline} for ZSD and GZSD because the former uses word vectors during training whereas the later does not adopt semantics. By contrast, in GZSD-seen detection cases, \method{Baseline} outperforms \method{Our-FL-word} because the use of unsupervised semantics (word vectors) during training in \method{Our-FL-word} introduces noise in the network which degrades the seen mAP. \textbf{(2)} From \method{Our-FL-word} to \method{Our-PL-word} unseen mAP improves because of the proposed loss which increases inter-class and reduces intra-class differences. It brings better visual-semantic alignment than FL (Fig.~\ref{fig:alignment_vis}). \textbf{(3)} \method{Our-PL-vocab} further improves the ZSD performance. Here, the vocabulary metric helps the word vectors to update based on visual similarity and allows features to align better with semantics.

\textbf{Semantics in Box-regression Subnet:} Our framework can be trained without semantics in the box-regression subnet. In Fig. \ref{fig:ablation} \textit{(Right)}, we compare performance with and without using word vectors in the regression branch using FL and our loss. We observe that using word vectors in regression branch helps to improve the performance of ZSD.

\textbf{Alternative formulation:} In the first row of Table \ref{tab:softvssigm}, we report performance of \texttt{Our-PL-vocab} using this alternative polarity loss with $\beta'=5$ and word2vec as semantic information. With this alternative formulation, we achieve a performance quite close to that of sigmoid-based polarity loss.

\begin{table*}[!htp]
\begin{center}
\begin{tabular}{|c|c|c|c|c|c|c|}
\hline
\multirow{3}{*}{Method} & \multirow{3}{*}{Seen/Unseen}&Word& \multirow{2}{*}{ZSD} & \multicolumn{3}{|c|}{GZSD}\\ \cline{5-7}
    	&&Vector&  & seen	& unseen & HM \\
&&  & (mAP) & (mAP)	& (mAP) & (mAP) \\ \hline
\texttt{Our-FL-vocab} &48/17&ftx&5.68&34.32&2.23&4.19 \\
\texttt{Our-PL-vocab}&48/17&ftx&6.99&35.13&2.73&5.07\\ \hline
\texttt{Our-FL-vocab} &65/15&glo&10.36&36.69&10.33&16.12\\
\texttt{Our-PL-vocab} &65/15&glo&{11.55}&{36.79}&{11.53}&{17.56} \\ \hline
\texttt{Our-FL-vocab}& 65/15 & w2v & 12.04 & \textbf{37.31}& 12.05 & 18.22 \\ 
\texttt{Our-PL-vocab}&65/15 & w2v & \textbf{12.62} & 32.99	& \textbf{12.62} & \textbf{18.26} \\ \hline
\end{tabular}
  \end{center}
\caption{\small More results on ZSD with different word-vectors (GloVe, FasText and Word2Vec).}
\label{tab:diffword}
\end{table*}

\begin{table*}[!htp]
\begin{center}
\begin{tabular}{|c|c|c|c|c|c|c|c|}
\hline
\multirow{3}{*}{Method} & \multirow{3}{*}{Penalty Function} & \multirow{3}{*}{Seen/Unseen}&Word& \multirow{2}{*}{ZSD} & \multicolumn{3}{|c|}{GZSD}\\ \cline{6-8}
    & 	&&Vector&  & seen	& unseen & HM \\
&& &  & (mAP) & (mAP)	& (mAP) & (mAP) \\ \hline
\texttt{Our-PL-vocab} &  softplus & 65/15 & w2v & 12.17 & 32.12 & 12.18 & 17.66 \\ \hline \hline
\texttt{Our-PL-vocab} & sigmoid &  65/15 & w2v & {12.62} & 32.99	& {12.62} & {18.26} \\ \hline
\end{tabular}
  \end{center}
\caption{\small Comparison of ZSD performance with softplus and sigmoid based penalty functions.}
\label{tab:softvssigm}
\end{table*}

\textbf{Choice of Semantic Representation:} In addition to word2vec as semantic word vectors reported in the main paper, we also experiment with GloVe \cite{Jeffrey_Glove_2014} and FastText \cite{fasttext_2017} as word vectors. We report those performances in Table \ref{tab:diffword}. We notice that Glove (glo) and FastText (ftx) achieve respectable performance, although they do not work as well as word2vec. However, in all cases, \texttt{Our-PL-vocab} beats \texttt{Our-FL-vocab} on ZSD in both cases.

\begin{figure}[!t]
  \centering
   \includegraphics[width=.5\textwidth,trim={.8cm .05cm 1.2cm .2cm},clip]{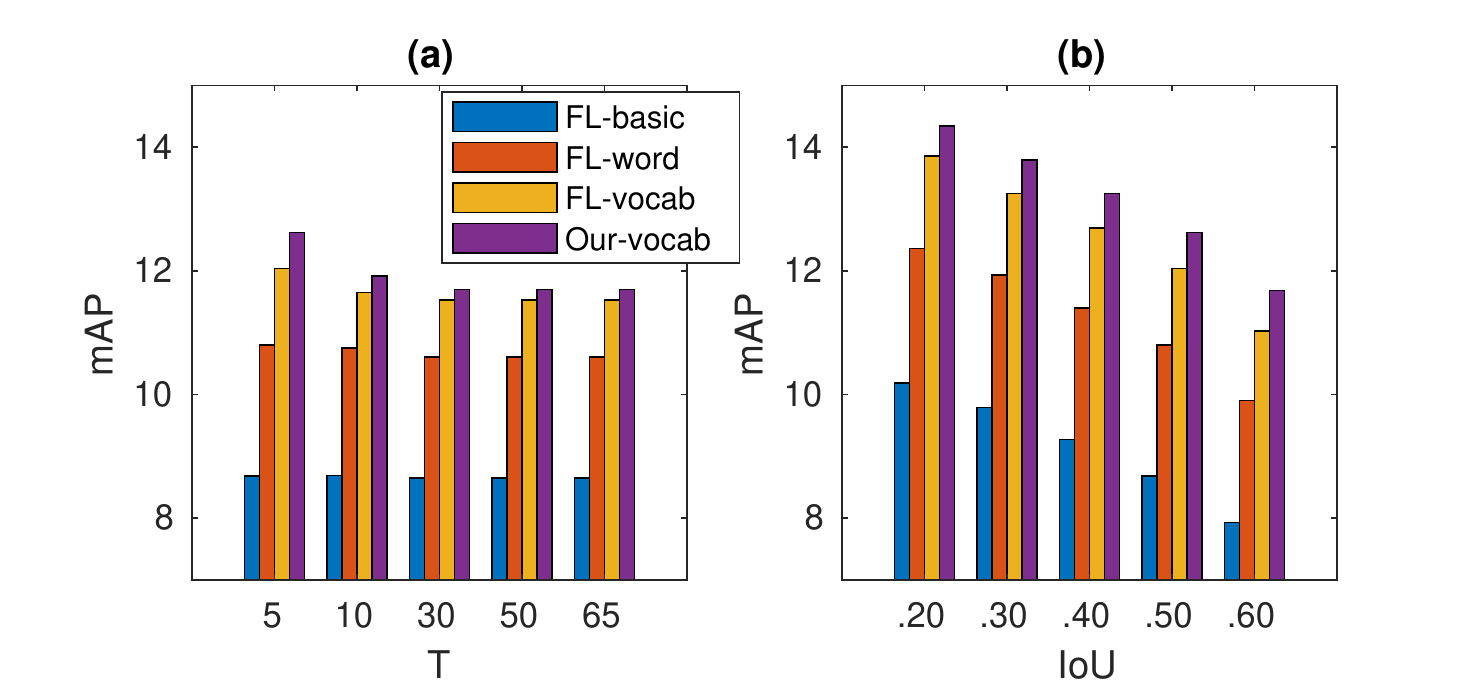}
   \caption{\small (a) Impact of selecting close seen (b) Impact of IoU.}
\label{fig:varyTandIoU}
\end{figure}

\textbf{Varying T and IoU:} In Sec.~5 of the main paper, we discussed the reduced description of an unseen class based on closely related seen classes. We experiment with the behavior of our model by varying a different number of close seen classes. One can notice in Fig.~\ref{fig:varyTandIoU}(a), a smaller number of close seen classes (e.g., 5) results in relatively better performance than using more seen classes (e.g., 10$-$65) during ZSD prediction. This behavior is related with the average number of classes per superclass (which is 7.2 for MS-COCO excluding `person') because dissimilar classes from a different superclass may not contribute towards describing a particular unseen class. Thus, we use only five close seen classes to describe an unseen in all our experiments. In Fig.~\ref{fig:varyTandIoU}(b), we report the impact of choosing a different IoU ratio (from $0.2$ to $0.6$) in ZSD. As expected, lower IoUs result in better performance than higher ones. As practiced in object detection literature, we use IoU$=0.5$ for all other experiments in this paper.

\begin{figure}[!t]
  \centering
   \includegraphics[width=.5\textwidth,trim={0cm 0cm 0cm 0cm},clip]{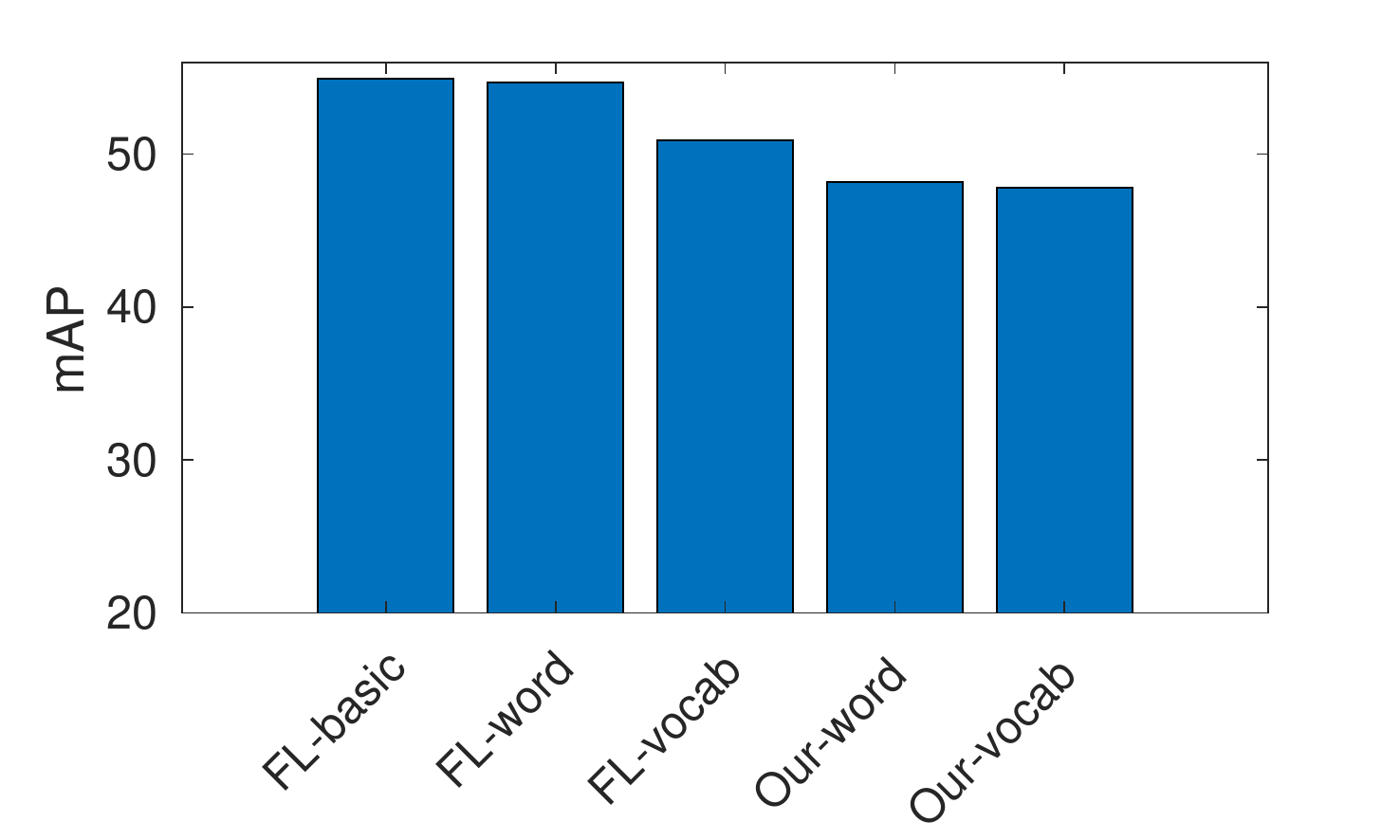}
   \caption{\small Traditional detection performance on 65 seen classes of MSCOCO.}
\label{fig:traditionadetection}
\end{figure}

{\textbf{Traditional detection:}} We report traditional detection performance of different versions of our framework in Fig.~\ref{fig:traditionadetection}. As a general observation, it is clear that making detection explicitly dependent on semantic information hurts the detector's performance on `seen' classes (traditional detection). This is consistent with the common belief that training directly on the desired output space in an end-to-end supervised setting achieves a better performance \cite{bojarski2016end,schmidhuber2015deep}. Consistently, we notice that \texttt{FL-basic} achieves the best performance because it is free from the noisy word vectors. \texttt{Our-FL-word} performs relatively worse than  \texttt{FL-basic} because of using noise word vectors as class semantics inside the network. Then, word vectors of vocabulary texts further reduce the performance in \texttt{Our-FL-vocab}. Our proposed loss (\texttt{Our-PL-word} and \texttt{Our-PL-vocab} cases) aligns visual features to noisy word vectors better than FL which is valuable for zero-shot learning but slightly degrades the seen performance. 
Similarly, we notice that while modifying the word embeddings, the vocabulary metric focuses more on proper visual-semantic alignment that is very helpful for ZSD but performs lower for the seen/traditional detection setting. 

\begin{figure*}[!t]
  \centering
   \includegraphics[width=1\textwidth,trim={0cm 0cm 0cm 0cm},clip]{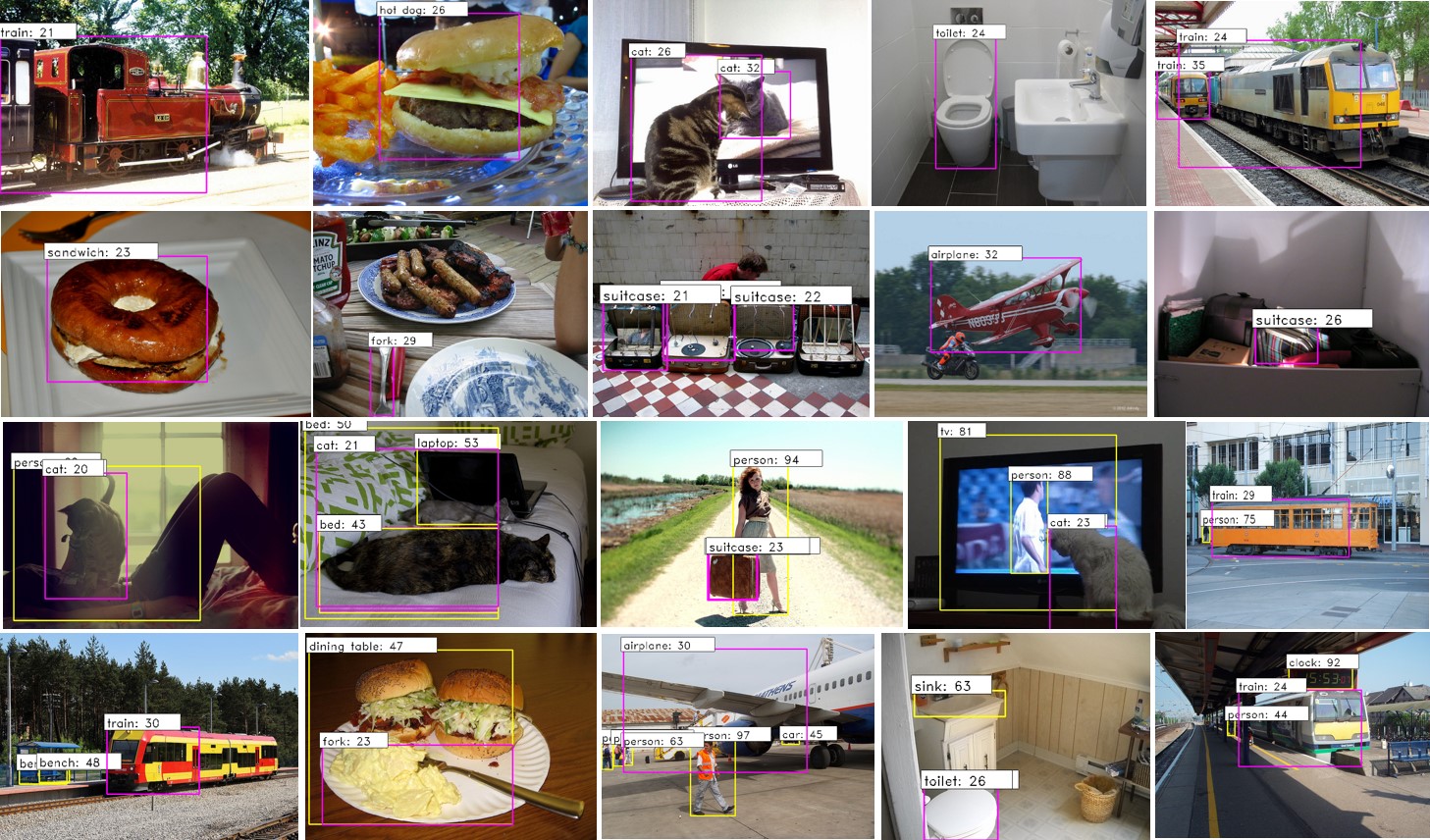}
   \caption{\small Qualitative results of ZSD (two top rows) and GZSD (two bottom rows). Pink and yellow boxes represent unseen and seen detections respectively.}
\label{fig:zsd2}
\end{figure*}

\textbf{Pascal VOC results:} To compare with  YOLO-based ZSD Demirel \emph{et al.}\cite{Demirel_BMVC_2018}, we adopt their exact settings with Pascal VOC 2007 and 2012. Note that, their approach used attribute vectors as semantics from \cite{aPY_2009}. As such attribute vectors are not available for our vocabulary list, we compare this approach with only using fixed attribute vectors inside our network. Our method beats Demirel \emph{et al.}\cite{Demirel_BMVC_2018} by a large margin ($57.9$ vs $63.5$ on traditional detection, $54.5$ vs $62.1$ on unseen detection). 


\section{Conclusion}

In this paper, 
we propose an end-to-end trainable framework for ZSD based on a new loss formulation. Our proposed polarity loss promotes correct alignment between visual and semantic domains via max-margin constraints and by dynamically refining the noisy semantic representations. Further, it penalizes an example considering background vs. object imbalance, easy vs. hard cases and inter-class vs. intra-class relations. We qualitatively demonstrate that in the learned semantic embedding space, word vectors become well-distributed and visually similar classes reside close together. We propose a realistic seen-unseen split on the MS-COCO dataset to evaluate ZSD methods. In our experiments, we have outperformed several recent state-of-the-art methods on both ZSD and GZSD tasks across the MS-COCO and Pascal VOC 2007 datasets.



\ifCLASSOPTIONcaptionsoff
  \newpage
\fi

\bibliographystyle{ieeetr}
\bibliography{ref}





\end{document}